\documentclass[10pt,twocolumn,letterpaper]{article}

\usepackage[pagenumbers]{iccv}
\usepackage{multirow}
\usepackage{stfloats}
\usepackage{colortbl}
\usepackage{makecell}
\usepackage{siunitx}
\usepackage[utf8]{inputenc} % allow utf-8 input
\usepackage[T1]{fontenc}    % use 8-bit T1 fonts
\usepackage{booktabs}       % professional-quality tables
\usepackage{amsfonts}       % blackboard math symbols
\usepackage{nicefrac}       % compact symbols for 1/2, etc.
\usepackage{microtype}      % microtypography
\usepackage{xcolor}         % colors
\usepackage{amsmath}
\usepackage{amsfonts,amssymb,amsthm} 
\usepackage{algorithm}        % 主要用于算法排版
\usepackage{algorithmic}      % 搭配 `algorithm` 使用，用于算法步骤
\usepackage{mathtools}        % 增强数学环境
\usepackage[pagebackref,breaklinks,colorlinks]{hyperref}
\definecolor{iccvblue}{rgb}{0.21,0.49,0.74}
\def\paperID{10193} % *** Enter the Paper ID here
\def\confName{ICCV}
\def\confYear{2025}

\newtheorem{theorem}{Theorem}

\title{Evolution-based Region Adversarial Prompt Learning for Robustness Enhancement in Vision-Language Models}

\author{Xiaojun Jia$^{1}$\thanks{The first two authors contribute equally to this work. }, Sensen Gao$^{3,*}$, Simeng Qin$^{2}$, Ke Ma$^{4}$, Xinfeng Li$^{1}$, \\ Yihao Huang$^{1}$, Wei Dong$^{1}$, Yang Liu$^{1}$, Xiaochun Cao$^{5}$ \\
$^{1}$Nanyang Technological University, Singapore \quad
$^{2}$Northeastern University, Shenyang, China\\
$^{3}$ Mohamed Bin Zayed University of Artificial Intelligence  \\
$^{4}$University of Chinese Academy of Sciences, Beijing, China\\
$^{5}$ School of Cyber Science and Technology, Shenzhen Campus of Sun Yat-sen University, China \\
{\tt\small \{jiaxiaojunqaq, sensen.gao2002, huangyihao22, lxfmakeit\}@gmail.com; qinsm@stumail.ysu.edu.cn;} \\ {\tt\small make@ucas.ac.cn; \{wei\_dong, yangliu\}@ntu.edu.sg; caoxiaochun@mail.sysu.edu.cn}
}

\begin{document}
\maketitle
\begin{abstract}
Large pre-trained vision-language models (VLMs), such as CLIP, demonstrate impressive generalization but remain highly vulnerable to adversarial examples (AEs). Previous work has explored robust text prompts through adversarial training, achieving some improvement in both robustness and generalization. However, they primarily rely on single-gradient direction perturbations (e.g., PGD) to generate AEs, which lack diversity, resulting in limited improvement in adversarial robustness. To address these limitations, we propose an evolution-based region adversarial prompt tuning method called ER-APT, which combines gradient methods with genetic evolution to generate more diverse and challenging AEs. In each training iteration, we first generate AEs using traditional gradient-based methods. Subsequently, a genetic evolution mechanism incorporating selection, mutation, and crossover is applied to optimize the AEs, ensuring a broader and more aggressive perturbation distribution. The final evolved AEs are used for prompt tuning, achieving region-based adversarial optimization instead of conventional single-point adversarial prompt tuning. We also propose a dynamic loss weighting method to adjust prompt learning efficiency for accuracy and robustness. Experimental evaluations on various benchmark datasets demonstrate the superiority of our proposed method, outperforming state-of-the-art APT methods. The code is released at \href{https://github.com/jiaxiaojunQAQ/ER-APT}{here}.

\end{abstract}    
\section{Introduction}
\begin{figure}[t]
    \centering
    \includegraphics[width=\linewidth]{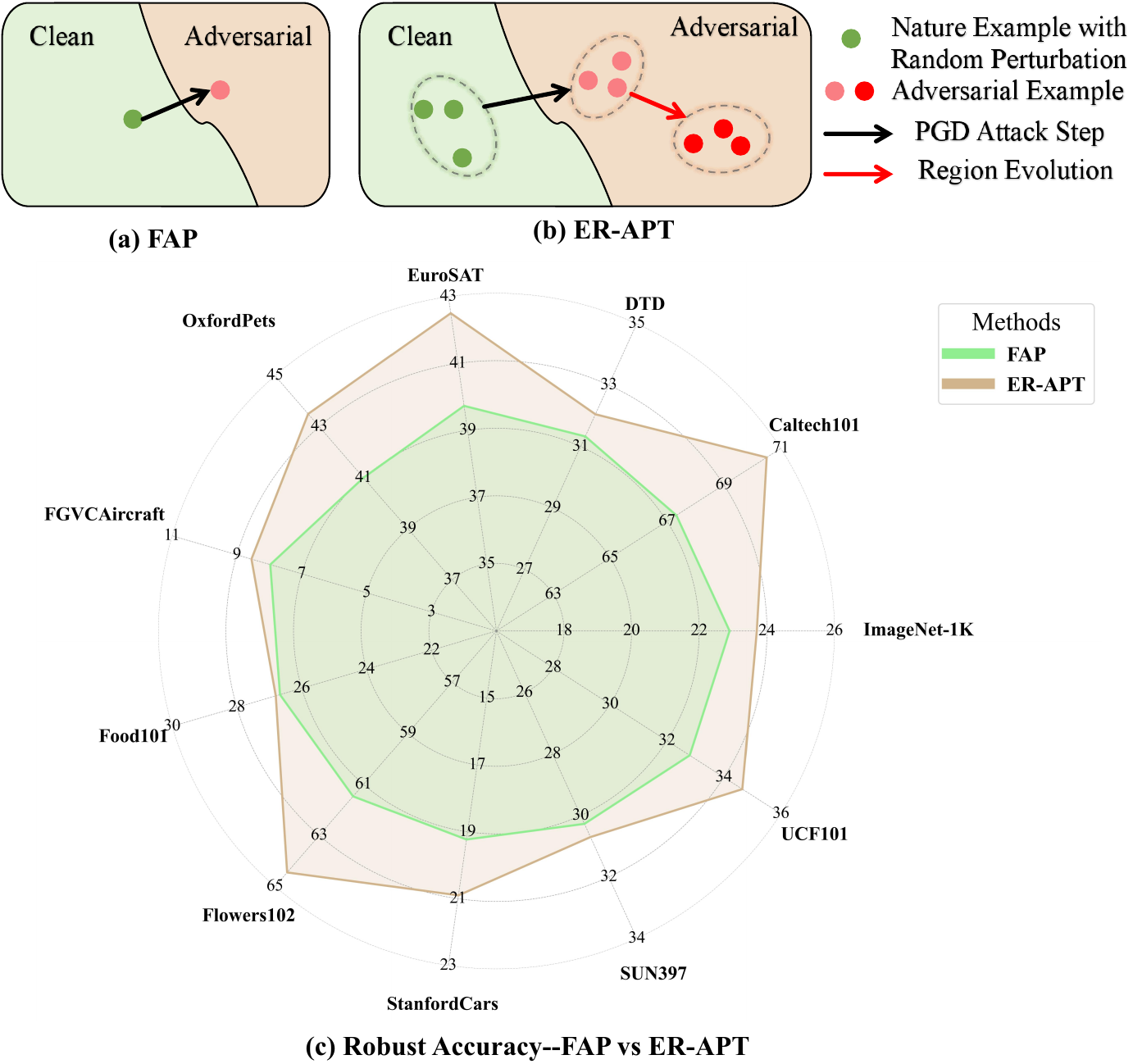}
      \caption{\textbf{FAP vs ER-APT}. (a) illustrates the adversarial examples generation in FAP~\cite{zhou2024fewshot}, which adopts the single-gradient direction to generate adversarial examples. (b) illustrates the generation of adversarial examples in our ER-APT, which combines gradient-based adversarial example generation with genetic evolution. (c) compares robustness of our ER-APT against FAP. }
    
    \label{fig:home}
\end{figure}

\label{sec:intro}
Large pre-trained Vision-Language Models (VLMs)~\cite{jia2021scaling,li2022blip,li2023blip}, which are trained on massive amounts of visual and language data, have achieved significant success across various domains of artificial intelligence such as image captioning~\cite{zhu2023prompt}, visual question answering~\cite{xie2024knowledge}, text-to-image generation~\cite{li2024safegen,zhang2023adding}. More and more studies~\cite{zhang2024vision,huang2025cutcp} have explored integrating VLMs into various downstream tasks, driven by the continuous emergence of new VLM architectures. However, as more research and applications~\cite{duan2024cityllava,saha2024improved} have been built on VLMs, any potential vulnerabilities could have a widespread impact on the performance and reliability of downstream tasks. More and more works~\cite{schlarmann2023adversarial,zhao2024evaluating,gao2025boosting,jia2024semantic} have focused on the vulnerability of VLMs and found that VLMs are vulnerable to adversarial examples (AEs)~\cite{szegedy2013intriguing,jia2020adv,gu2023survey,huang2023ALA,huang2024TSCUAP,huang2024perception,huang2025scale}, that is, the model output can be manipulated by imperceptible adversarial perturbations on the image~\cite{shirnin2024analyzing,zhang2024avibench}. It poses significant safety risks and raises serious concerns about their reliability and security.

\par To improve the adversarial robustness against adversarial attacks, a series of works~\cite{jia2019comdefend,zhang2021defense,huang2024adversarial,li2024vrifle} have been proposed. Among them, adversarial training~\cite{pang2020bag,jia2022adversarial,jia2024improving} has been one of the most effective and widely used methods, which can be formulated as a mini-max problem to improve adversarial robustness. Enhancing the robustness of large-scale VLMs with a large number of parameters through adversarial training from scratch is impractical, as the process requires significant computational resources for generating AEs at each training step. Fine-tuning~\cite{yosinski2014transferable} offers an efficient approach to adapting pre-trained models for downstream tasks. However, as these models grow to tens or even hundreds of billions of parameters, updating all their weights becomes increasingly costly. The expense increases even further when adversarial training is introduced to improve adversarial robustness.
 
\par To further improve the efficiency of training, prompt tuning~\cite{zhou2022learning, jia2022visual, khattak2023maple, yao2023visual} has emerged as a more efficient alternative to traditional fine-tuning, enabling pre-trained models to adapt to downstream tasks. Instead of updating the model’s weights, this method introduces learnable prompt vectors, allowing the model to transfer knowledge effectively. When incorporating adversarial training for fine-tuning VLMs, prompt learning not only boosts adversarial robustness but also improves training efficiency. Recent works~\cite{chen2023visual,li2024one,zhang2024adversarial,zhou2024fewshot,yang2024revisiting,yuan2025promptguard} have leveraged the prompt tuning to improve the adversarial robustness of VLMs in downstream tasks. For example, \citet{chen2023visual} propose to adopt adversarial visual prompting~\cite{bahng2022exploring} to boost the adversarial robustness of VLPs at test time. \citet{li2024one} indicate that CLIP’s robustness against adversarial attacks depends significantly on the choice of inference prompts and propose to combine text prompt tuning~\cite{zhou2022conditional,zhou2022learning} with adversarial to improve the adversarial robustness. \citet{zhang2024adversarial} propose to integrate multi-modal prompt~\cite{khattak2023maple} learning into adversarial training to enhance the robustness against adversarial attacks, which achieves the best robustness performance. 
\par Although they have achieved significant improvement in adversarial robustness, they only consider using the single-gradient direction to generate the AEs for training, resulting in limited robustness improvement. First, VLMs~\cite{radford2021learning,zhou2023non} have highly nonlinear characteristics, leading to local nonlinearity of the loss function. During each attack iteration, the adversarial samples generated based on a single gradient may not reach the local maximum region of the loss function, resulting in low-quality adversarial samples and limited improvement in adversarial robustness. Second, previous works primarily rely on single-gradient direction perturbations (e.g., PGD) to generate AEs. However, these methods often lead to overfitting specific attack patterns and struggle to generalize against diverse or adaptive attacks. 

\par To address these issues, we propose an evolution-based region few-shot adversarial prompt tuning method called ER-APT. The proposed method integrates gradient-based optimization with genetic evolution to enhance the adversarial robustness. The ER-APT framework consists of two primary stages: gradient-based adversarial example generation and evolutionary refinement. In the first stage, AEs are generated using gradient-based methods, targeting the most vulnerable regions of the space to induce misalignment in model outputs. In the second stage, a genetic algorithm is employed to iteratively refine these AEs through selection, mutation, and crossover operations. This evolutionary mechanism facilitates the discovery of more diverse and high-quality AEs for few-shot prompt tuning to effectively improve the adversarial robustness. Unlike conventional adversarial prompt tuning techniques that focus on single-point perturbations, ER-APT employs a region-based optimization strategy, enabling broader and more comprehensive adversarial space exploration. 
\par Moreover, previous studies indicate that there is a trade-off between clean accuracy and adversarial robustness. We also propose a dynamic loss weighting method to balance clean accuracy and adversarial robustness. Specifically, we fine-tune the prompts of VLMs on the clean and adversarial samples and regard the improvement of clean accuracy and adversarial robustness as the two tasks. The convergence rate of a task can be represented by its learning speed—the faster the task loss decreases, the higher the learning speed. As the learning speed increases, the task loss is assigned a lower weight.  

 % dynamically adjusting the loss weights for clean and adversarial samples in response to the magnitude of their gradients. 
\par We conduct a series of experiments to evaluate the effectiveness of our ER-APT across various network architectures and datasets. The experimental results indicate that the proposed method can significantly improve the adversarial robustness of VLMs, surpassing the state-of-the-art APT. Our main contributions are in three aspects:
\begin{enumerate}[itemsep=2pt,topsep=0pt,parsep=0pt]
    \item We propose an evolution-based region few-shot adversarial prompt tuning method (ER-APT) that combines gradient-based optimization with genetic evolution to enhance adversarial robustness.
    \item We propose a dynamic loss weighting method that balances clean accuracy and adversarial robustness by adjusting loss weights based on the converged task speed, assigning lower weights to faster-converging tasks.
    \item Extensive experiments across various network architectures and datasets demonstrate the superiority of our ER-APT, outperforming state-of-the-art adversarial prompt tuning methods.
\end{enumerate}
\section{Related Work}
\label{sec:related_work}

\subsection{Prompt tuning for VLMs}
Prompt tuning~\cite{zhou2022conditional,zhou2022learning,khattak2023maple} has emerged as an effective method for adapting large pre-trained Vision-Language Models (VLMs) like CLIP to downstream tasks with minimal computational cost. Instead of updating all model parameters, prompt tuning optimizes a small set of learnable parameters, making it a parameter-efficient alternative to full fine-tuning. \citet{zhou2022learning} introduce the concept of learnable textual prompts as a replacement for hand-crafted prompts to enhance textual embeddings, a technique known as context optimization (CoOp). Building upon this, \citet{zhou2022conditional} addresses the overfitting issue in CoOp by conditioning the prompts on visual features, leading to improved performance. This approach is referred to as conditional context optimization (CoCoOp). \citet{chen2022plot} introduce the use of optimal transport to align vision and text modalities, enabling the generation of discriminative and visually coherent local textual prompts. Expanding beyond textual prompt tuning, \citet{khattak2023maple} propose Multi-modal Prompt Learning (MaPLe), which incorporates prompt tuning in both the vision and language branches to enhance the alignment between visual and textual representations.

\subsection{Adversarial Prompt Tuning in VLMs}
Some works have indicated that VLMs are vulnerable to AEs, which are generated by adding some adversarial perturbations. Many defense methods have been proposed to improve adversarial robustness, among which adversarial training stands out. Recent works combine prompt tuning with adversarial training to efficiently improve adversarial robustness. Specifically, \citet{chen2023visual} propose Class-wise Adversarial Visual Prompting (C-AVP), which enhances adversarial robustness by generating class-specific visual prompts, outperforming conventional methods in both standard and robust accuracy. \citet{li2024one} proposes Adversarial Prompt Tuning (APT) to enhance the adversarial robustness of VLMs by learning robust text prompts instead of modifying model weights. \citet{zhang2024adversarial} first adopt the image encoder to generate AEs and leverage learnable text prompts aligned with the generated adversarial image embeddings to improve the adversarial robustness of VLMs, called Adversarial Prompt Tuning (AdvPT). \citet{zhou2024fewshot} propose leveraging multi-modal prompts~\cite{khattak2023maple} for adversarial training and designing a loss function that balances the connection between natural and adversarial features across different modalities, thereby improving adversarial robustness.

\section{The Proposed Method}

% You must include your signed IEEE copyright release form when you submit your finished paper.
% We MUST have this form before your paper can be published in the proceedings.

% Please direct any questions to the production editor in charge of these proceedings at the IEEE Computer Society Press:
% \url{https://www.computer.org/about/contact}.
\begin{figure*}[t]
    \centering
    \includegraphics[width=\linewidth]{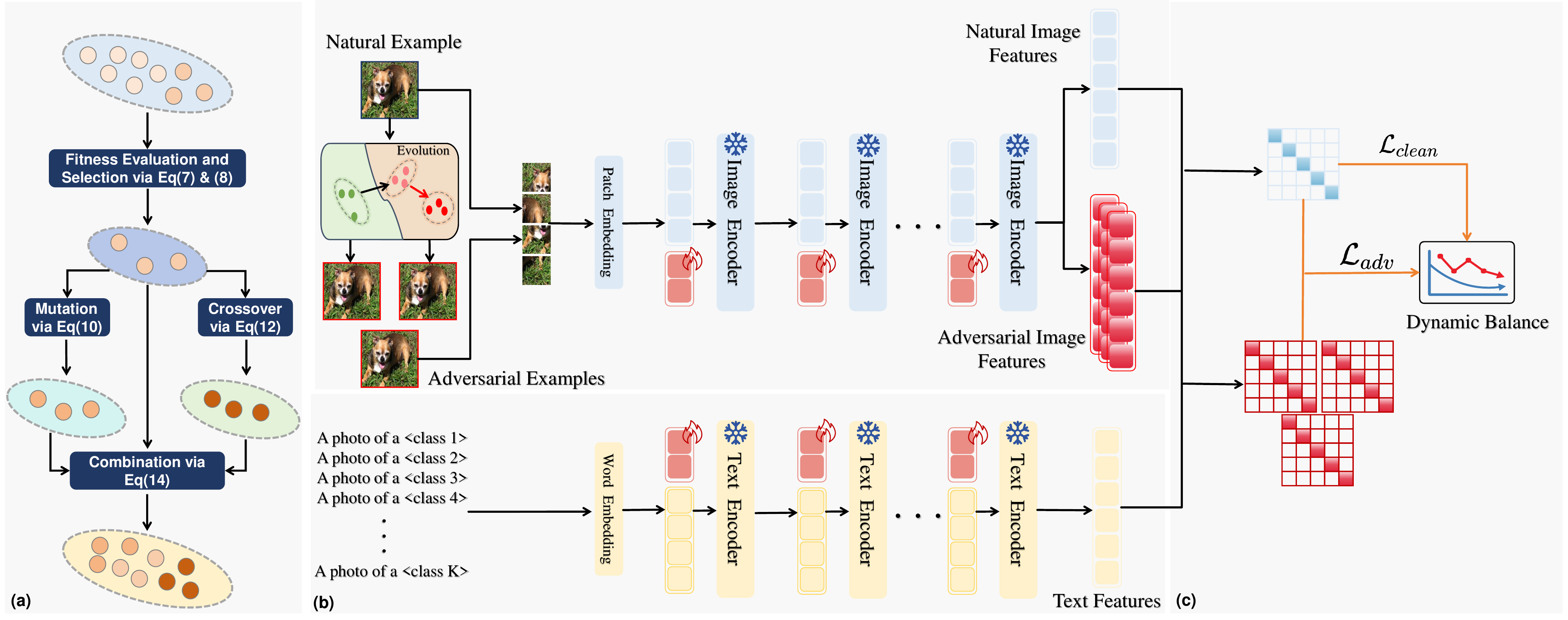}
    \vspace{-6mm}
    \caption{\textbf{The Pipeline of the Proposed ER-APT.} (a) Pipeline for the evolution-based region in adversarial example generation. (b) Pipeline for the generation of the text and image features. (c) Pipeline for the dynamic loss weighting.}
    \label{fig:method}
\end{figure*}

\subsection{Preliminary Study}
CLIP comprises two key components: an image encoder $f_{I}(\cdot)$ with the parameters $\mathcal{W}_{v}$ and a text encoder $f_{T}(\cdot)$ with the parameters $\mathcal{W}_{t}$. Given a $N$-class image classification task with an input image $\boldsymbol{x}$ and the corresponding label $y \in\{1, \ldots, N\}$, we divide the image $\boldsymbol{x}$ into $K$ patches and embed them as: $\boldsymbol{h}(\boldsymbol{x})=\left\{c_{\mathrm{cls}}, h_1(\mathbf{x}), \ldots, h_K(\mathbf{x})\right\}$, where $c_{\mathrm{cls}}$ is a class token and $h_i(\mathbf{x})$ is the $i$-th image patch embedding. The image features can be extracted from the patch embedding and represented as: $\boldsymbol{z}_v=f_{I}\left(\boldsymbol{h}(\boldsymbol{x}) ; \mathcal{W}_{v}\right)$.  Then  we generate handcrafted prompts $t_j \in \boldsymbol{t}=\left\{t_1, \ldots, t_N\right\}$ by inserting the class name into a template like ``a photo of a {class}.'' Each prompt is then tokenized and embedded as: $\boldsymbol{d}\left(t_j\right)=\left\{d_1\left(t_j\right), \ldots, d_M\left(t_j\right) \right\}$, where $d_i\left(t_j\right)$ is the $i$-th token text embedding. The text features can be extracted from the text embedding and represented as: $\boldsymbol{z}_t^j=f_{t}\left(\boldsymbol{d}\left(t_j\right) ; \mathcal{W}_{t}\right)$. 
% These similarity scores function similarly to the logit outputs in traditional classification models. The probability of $\boldsymbol{x}$ aligning with 
% $\boldsymbol{t}_{j}$ can be calculated as:
For the image classification, the probability of the image $\boldsymbol{x}$ in the $j$-th class is calculated as:
\begin{equation}
p(\boldsymbol{x},\boldsymbol{t}_j)=\frac{\exp \left(\cos \left(\boldsymbol{z}_v, \boldsymbol{z}_t^j\right) / \eta\right)}{\sum_{j=1}^K \exp \left(\cos \left(\boldsymbol{z}_v, \boldsymbol{z}_t^j\right) / \eta\right)},
\end{equation}
where $\cos(\cdot)$ represents the cosine similarity score, $K$ represents the total number of the given text, and $\eta$ represents the temperature parameter. 
\par Prompt learning aims to train lightweight, learnable prompts using only a few examples from downstream data. The prompt learning can be divided into text prompt learning and visual prompt learning. In terms of text prompt learning, the learnable text prompt $\boldsymbol{P}_t$ is integrated into text embeddings as: $\boldsymbol{d}\left(t_j, \boldsymbol{P}_t \right)=\left\{d_1\left(t_j\right), \ldots, d_M\left(t_j\right), \boldsymbol{P}_t \right\}$. The text features can represented as:  $\boldsymbol{z}_t^{(j,\boldsymbol{P}_t)}=f_{t}\left(\boldsymbol{d}\left(t_j, \boldsymbol{P}_t\right) ; \mathcal{W}_{t}\right)$. By leveraging cross-entropy, the learnable text prompts are optimized to reduce the distribution gap between text-image logits and the ground-truth label. It can be calculated as:
\begin{equation}
\boldsymbol{P}_{{t}}^*=\arg \min _{\boldsymbol{P}_{{t}}} \mathbb{E}_{(\boldsymbol{x}, y)} \mathcal{L}_{\mathrm{CE}}\left(\cos \left(\mathbf{z}_{v}, \mathbf{z}_t^{(j,\boldsymbol{P}_t)}\right), y\right). 
\end{equation}
In terms of visual prompt learning, the learnable visual prompt $\boldsymbol{P}_v$ is integrated into image embeddings as: $\boldsymbol{h}(\boldsymbol{x}, \boldsymbol{P}_v)=\left\{c_{\mathrm{cls}}, h_1(\mathbf{x}), \ldots, h_K(\mathbf{x}), \boldsymbol{P}_v\right\}$. The features of the image can be represented as:
$\boldsymbol{z}_{(v,\boldsymbol{P}_v)}=f_{I}\left(\boldsymbol{h}(\boldsymbol{x}, \boldsymbol{P}_v) ; \mathcal{W}_{v}\right)$. The learnable image prompt is optimized as:
\begin{equation}
\boldsymbol{P}_{{v}}^*=\arg \min _{\boldsymbol{P}_{{v}}} \mathbb{E}_{(\boldsymbol{x}, y)} \mathcal{L}_{\mathrm{CE}}\left(\cos \left(\mathbf{z}_{(v,\boldsymbol{P}_v)}, \mathbf{z}_t^{j}\right), y\right). 
\end{equation}

Hence, composed of both image and text prompts, the learnable prompt is represented as $\boldsymbol{P}=\left\{\boldsymbol{P}_{\boldsymbol{v}}, \boldsymbol{P}_{\boldsymbol{t}}\right\}$. The learnable prompt is optimized as:
\begin{equation}
\boldsymbol{P}^*=\arg \min _{\boldsymbol{P}} \mathbb{E}_{(\boldsymbol{x}, y)} \mathcal{L}_{\mathrm{CE}}\left(\cos \left(\mathbf{z}_{(v,\boldsymbol{P}_v)}, \mathbf{z}_t^{(j,\boldsymbol{P}_t)}\right), y\right),
\end{equation}
where $\cos \left(\mathbf{z}_{(v,\boldsymbol{P}_v)}, \mathbf{z}_t^{(j,\boldsymbol{P}_t)}\right)$ is the text-image logits, which can be abbreviated as $f_{\boldsymbol{P}}(\boldsymbol{x})$.

To boost robustness, the prompt learning combined with adversarial tuning can be formulated a mini-max problem:
\begin{equation}
\label{eq:sat}
\min _{\boldsymbol{P}} \mathbb{E}_{(\boldsymbol{x}, y) \sim \mathcal{D}}\left[\max _{\boldsymbol{\delta} \in \Omega} \mathcal{L}\left(f_{\boldsymbol{P}}(\boldsymbol{x}+\boldsymbol{\delta}), y\right)\right],
\end{equation}
where $\mathcal{D}$ is the data distribution, $\Omega$ is the set of the perturbation, and $\boldsymbol{\delta}$ is the adversarial perturbation. 

% They are used to extract the features from an input image \boldsymbol{x}_{i}\boldsymbol{x}_{i} and text \boldsymbol{t}_{j}\boldsymbol{t}_{j} respectively. The image features can be represented as \boldsymbol{z}_v^i=f_{I}\left(\boldsymbol{x}_i ; \mathcal{W}_{v}\right)\boldsymbol{z}_v^i=f_{I}\left(\boldsymbol{x}_i ; \mathcal{W}_{v}\right) and the text features can be represented as
% \boldsymbol{z}_t^j=f_{t}\left(\boldsymbol{t}_j ; \mathcal{W}_{t}\right)\boldsymbol{z}_t^j=f_{t}\left(\boldsymbol{t}_j ; \mathcal{W}_{t}\right). 

\subsection{Pipeline of The Proposed Method}
\label{sec:pipline} 
Previous works use a single-gradient direction to solve the internal maximization problem in Eq.~\ref{eq:sat} for generating adversarial examples (AEs), leading to limited robustness gains. To address this, we integrate gradient-based AE generation with genetic evolution to enhance AE quality and improve adversarial robustness, as illustrated in Fig.~\ref{fig:method}.
% Previous works adopt the single-gradient direction to solve the internal maximization problem in Eq.~\ref{eq:sat} for generating AEs, resulting in limited robustness improvement. To overcome this limitation,
% we propose to combine the gradient-based adversarial example generation with the genetic evolution to improve the quality of AEs, thereby improving adversarial robustness. 
% The pipeline of the proposed method is shown in Fig.~\ref{fig:method}. 

\noindent \textbf{Adversarial example initialization.} During the $t$-th attack iteration, given the adversarial example population
$\mathcal{X}^{t-1}=\left\{\boldsymbol{x}+\boldsymbol{\delta}_1^{t-1}, \boldsymbol{x}+\boldsymbol{\delta}_2^{t-1}, \ldots, \boldsymbol{x}+\boldsymbol{\delta}_\mathcal{N}^{t-1}\right\}$, where $\mathcal{N}$ is the initial population number. Then, we adopt the model gradients to generate initialized AEs, calculated as:
\begin{equation}
\boldsymbol{\delta}_i^{t}=\Pi_{[-\epsilon, \epsilon]} [\boldsymbol{\delta}_i^{t-1}+\alpha \operatorname{sign}(\nabla_{\boldsymbol{\delta}_i^{t-1}} \mathcal{L}(f_{\boldsymbol{P}}(\boldsymbol{x}+\boldsymbol{\delta}_i^{t-1}), y))],
\end{equation}
where $\Pi_{[-\epsilon, \epsilon]}$ represents the projection operator onto the set $[-\epsilon, \epsilon]$, $\alpha$ represents the step size and $\boldsymbol{x}_i^{t-1}$ represent the $i$-th sample in the population. 

\noindent \textbf{Fitness Evaluation and Selection.} To guide the evolutionary process, we define fitness using adversarial loss:
\begin{equation} 
\mathcal{F}(\boldsymbol{x}+\boldsymbol{\delta}_i^{t}) = \mathcal{L} \left( f_{\boldsymbol{P}} (\boldsymbol{x}+\boldsymbol{\delta}_i^{t}), y \right). 
\end{equation}
Higher fitness values indicate stronger adversarial effectiveness. We rank the population in descending order based on their fitness and retain the top $\mathcal{N}/3$ individuals:
\begin{equation}
\mathcal{X}^{\text {selected}, t}=\left\{\boldsymbol{x}+\boldsymbol{\delta}_{s_1}^t, \boldsymbol{x}+\boldsymbol{\delta}_{s_2}^t, \ldots, \boldsymbol{x}+\boldsymbol{\delta}_{s_{\mathcal{N} / 3}}^t\right\},
\end{equation}
where the indices $s_1, s_2, \ldots, s_{s_{\mathcal{N} / 3}}$ are sorted such that:
\begin{equation}
\mathcal{F}\left(\boldsymbol{x}+\boldsymbol{\delta}_{s_1}^t\right) \geq \mathcal{F}\left(\boldsymbol{x}+\boldsymbol{\delta}_{s_2}^t\right) \geq \cdots \geq \mathcal{F}\left(\boldsymbol{x}+\boldsymbol{\delta}^t_{s_{\mathcal{N} / 3}}\right)
\end{equation}
\noindent \textbf{Mutation.} Mutation introduces diversity into the population by adding small random perturbations to the selected individuals. Given an individual $\boldsymbol{x}_{s_i}$, the mutation process generates a perturbed variant as: 
\begin{equation}
\boldsymbol{\delta}^t_{m_i}=\Pi_{[-\epsilon, \epsilon]}[\boldsymbol{\delta}^t_{s_i} +\xi], \quad \xi \sim\mathcal{U}(-\phi \times \epsilon, \phi \times \epsilon),
\end{equation}
% TODO
where $\mathcal{U}(-\phi \times \epsilon, \phi \times \epsilon)$ represents a uniform random perturbation, with 
$\phi$ controlling the intensity of the mutation perturbation. The mutated individuals can be represented as:
\begin{equation}
\mathcal{X}^{\text {mutated}, t}=\left\{\boldsymbol{x}+\boldsymbol{\delta}_{m_1}^t, \boldsymbol{x}+\boldsymbol{\delta}_{m_2}^t, \ldots, \boldsymbol{x}+\boldsymbol{\delta}_{m_{\mathcal{N} / 3}}^t\right\},
\end{equation}

\noindent \textbf{Crossover.}
To further explore the adversarial space, we introduce crossover operations by linearly combining two randomly selected parents from the selected population:
\begin{equation}
\boldsymbol{\delta}^t_{c_{i}}=\Pi_{[-\epsilon, \epsilon]} [\lambda \boldsymbol{\delta}_{p_1}^t+(1-\lambda) \boldsymbol{\delta}_{p_2}^t],
\end{equation}
where $\boldsymbol{\delta}_{p_1}^t$, $\boldsymbol{\delta}_{p_2}^t$ are two distinct individuals randomly sampled from $\mathcal{X}^{\text {selected }, t}$, and $\lambda \sim \mathcal{U}(0,1)$ is a randomly sampled weighting coefficient. The crossover-generated individuals can be represented as:
\begin{equation}
\mathcal{X}^{\text {crossover}, t}=\left\{\boldsymbol{x}+\boldsymbol{\delta}_{c_1}^t, \boldsymbol{x}+\boldsymbol{\delta}_{c_2}^t, \ldots, \boldsymbol{x}+\boldsymbol{\delta}_{c_{\mathcal{N} / 3}}^t\right\},
\end{equation}
% TODO Sensen
Finally, the new population at iteration $t$ consists of both mutated and crossover-generated AEs:
 \begin{equation} \mathcal{X}^{t} = \mathcal{X}^{\text{mutated}, t} \cup \mathcal{X}^{\text{crossover}, t} \cup \mathcal{X}^{\text {selected}, t}. 
\end{equation}
Note that in the last generation, only selected samples $\mathcal{X}^{\text {selected}, t}$ are used for prompt tuning.
 % The final adversarial perturbations are projected back into the perturbation range. 

\subsection{Formulation of The Proposed Method}
Instead of directly optimizing a single perturbation $\boldsymbol{\delta}$, our method evolves a population of AEs iteratively. The inner maximization problem is formulated as:
\begin{equation}
\max _{\delta \in \mathcal{G}(\boldsymbol{x}, \mathcal{N})} \mathcal{L}\left(f_\theta(\boldsymbol{x}+\delta), y\right),
\end{equation}
where $\mathcal{G}(\boldsymbol{x}, \mathcal{N})$ is obtained through an iterative evolutionary process, which is presented in Sec.~\ref{sec:pipline}. The learnable prompt $\mathbf{P}$ is updated by minimizing the loss over the AEs from the evolved population: 
\begin{equation}
\min _\mathbf{P} \mathbb{E}_{(\boldsymbol{x}, y) \sim D}\left[\frac{1}{\mathcal{N}} \sum_{\delta_i \in \mathcal{G}(\boldsymbol{x}, \mathcal{N})} \mathcal{L}\left(f_\mathbf{P}\left(\boldsymbol{x}+\delta_i\right), y\right)\right].
\end{equation}
Hence, the mini-max formulation of the proposed method can be calculated as:
\begin{equation}
\label{eq:objective}
\min _\theta \mathbb{E}_{(\boldsymbol{x}, y) \sim D}\left[\max _{\delta_i \in \mathcal{G}(\boldsymbol{x}, \mathcal{N})} \frac{1}{\mathcal{N}} \sum_{i=1}^\mathcal{N} \mathcal{L}\left(f_\theta\left(\boldsymbol{x}+\delta_i\right),y\right)\right]. 
\end{equation}
Compared with standard adversarial training (Eq.~\ref{eq:sat}), the proposed method leverages evolutionary method to generate diverse AEs, overcoming the limitation of standard adversarial training, which optimizes against only a single worst-case perturbation. By incorporating fitness-based selection, mutation, and crossover, our method explores a broader range of adversarial perturbations within the attack space, leading to better adversarial generalization and enhanced robustness. 

\subsection{Proposed Dynamic Loss Weighting Method}
Previous works indicate that there is a trade-off between clean accuracy and adversarial robustness. Following \cite{zhang2019theoretically,zhou2024fewshot}, we adopt both accuracy and robustness losses to perform adversarial prompt learning, calculated as: 
\begin{equation}
\label{eq:loss}
\mathcal{L}=\alpha \mathcal{L}_{\text{CE}}\left(f_{\boldsymbol{P}}(\boldsymbol{x}), y\right) + \beta \mathcal{L}_{\text{KL}}\left(f_{\boldsymbol{P}}(\boldsymbol{x}), f_{\boldsymbol{P}}(\boldsymbol{x}+\boldsymbol{\delta}) \right),
\end{equation}
where $\mathcal{L}_{\text{CE}}$ represents the the cross entropy loss between the clean image logits and ground truth label,  $\mathcal{L}_{\text{KL}}$ represents the Kullback–Leibler (KL) divergence between the logits of clean images and adversarial examples, and $\alpha$ and $\beta$ represents the weight parameters. Previous works manually set $\alpha$ to 1.0 and $\beta$ to 1.5 to conduct adversarial prompt learning. 
However, these methods require significant domain expertise, and their improvements in robustness remain limited. They rely on hand-crafted loss weight strategies that are independent of model states, without leveraging any state-specific information. Since models exhibit statistical variations in robustness throughout training, loss weight strategies should be designed adaptively based on the model state information. To further boost adversaria robustness, we propose a dynamic loss weighting method to balance model accuracy and robustness. Specifically, we fine-tune the prompts of VLMs using both clean and adversarial samples, treating improvements in clean accuracy and adversarial robustness as two distinct tasks. The convergence rate of each task is measured by its learning speed—the faster the task loss decreases, the higher the learning speed. At the  $\mathbb{T}$-th time, learning speed can be  calculated by the ratio of the loss at epoch $\mathbb{T}$ and epoch $\mathbb{T}-1$:
\begin{equation}
\begin{split}    
w_{\text{acc}}(\mathbb{T}) &=\frac{\mathcal{L}_{\text{CE}}^{\mathbb{T}}\left(f_{\boldsymbol{P}}(\boldsymbol{x}), y\right)}{\mathcal{L}_{\text{CE}}^{\mathbb{T}-1}\left(f_{\boldsymbol{P}}(\boldsymbol{x}), y\right)}, \\
w_{\text{rob}}(\mathbb{T})&=\frac{\mathcal{L}_{\text{KL}}^{\mathbb{T}}\left(f_{\boldsymbol{P}}(\boldsymbol{x}), f_{\boldsymbol{P}}(\boldsymbol{x}+\boldsymbol{\delta}) \right)}{\mathcal{L}_{\text{KL}}^{\mathbb{T}-1}\left(f_{\boldsymbol{P}}(\boldsymbol{x}), f_{\boldsymbol{P}}(\boldsymbol{x}+\boldsymbol{\delta}) \right)},
\end{split}
\end{equation}
where $w_{\text{acc}}(\mathbb{T})$ represents the learning speed of the clean accuracy and $w_{\text{rob}}(\mathbb{T})$represents the learning speed of the adversarial robustness at the time $\mathbb{T}$. The weight parameters in Eq.~\ref{eq:loss} can be calculated by: 
\begin{equation}
\begin{split}  
\alpha & = \alpha_{init} \times 2.0 \times \frac{\exp \left(w_{\text{acc}}(\mathbb{T}) / T\right)}{\exp \left(w_{\text{acc}}(\mathbb{T}) / T\right)+ \exp \left(w_{\text{rob}}(\mathbb{T}) / T\right)}, \\ 
\beta & = \beta_{init} \times 2.0 \times \frac{\exp \left(w_{\text{rob}}(\mathbb{T}) / T\right)}{\exp \left(w_{\text{acc}}(\mathbb{T}) / T\right)+ \exp \left(w_{\text{rob}}(\mathbb{T}) / T\right)},
\end{split}
\end{equation}
where $\alpha_{init}$ and $\beta_{init}$ denote the initial values of $\alpha$ and $\beta$ at the beginning of training, following the FAP setting of $1.0$ and $1.5$. The multiplication by $2.0$ ensures fluctuations around $1.0$, adjusting the initialization of $\alpha,\beta$. Additionally, $T$ represents the temperature coefficient, which modulates the weight differences among tasks. The detailed algorithm is provided in the Appendix. 

\subsection{Theoretical Analysis}
\begin{theorem}
    Suppose that the loss function $\mathcal{L}$ is $L$-Lipschitz in $\boldsymbol{P}$, \textit{i.e.}
    \begin{equation*}
        |\mathcal{L}(\boldsymbol{P},\boldsymbol{x}+\delta)-\mathcal{L}(\boldsymbol{P},\boldsymbol{x}+\delta')|\leq L\|\delta-\delta'\|,\ \ \forall\ \delta,\delta'\in\boldsymbol{\Delta}_{\epsilon},
    \end{equation*}
    where $\boldsymbol{\Delta}_{\epsilon}$ is the allowable perturbation space as
    \begin{equation}
        \boldsymbol{\Delta}_{\epsilon} = \{\delta\in\mathbb{R}^d:\ \|\delta\|\leq\epsilon\}.
    \end{equation}
    If ER-APT \eqref{eq:objective} optimizes the learnable prompts $\boldsymbol{P}$ to satisfy
    \begin{equation}
        \frac{1}{\mathcal{N}}\sum_{i=1}^{\mathcal{N}}\mathcal{L}(\boldsymbol{P},\boldsymbol{x}+\delta_i)\leq \gamma,
    \end{equation}
    then for any perturbation $\delta\in\boldsymbol{\Delta}_{\epsilon}$, we have
    \begin{equation}
        \mathcal{L}(\boldsymbol{P},\boldsymbol{x}+\delta)\leq \gamma+L\eta,
    \end{equation}
    where $\eta$ satisfies there exists $\delta_i\in\mathcal{P}$ such that $\|\delta-\delta_i\|\leq\eta$, and $\mathcal{P} = \{\delta_1,\delta_2,\dots,\delta_N\}$ denote the population of perturbations maintained by ER-APT. 
\end{theorem}
\noindent The evolutionary mechanism in ER-APT explicitly enforces diversity in $\mathcal{P}$ (via mutation and crossover), ensuring $\eta$ is small. This tightens the bound $\gamma+L\eta$, improving robustness across $\boldsymbol{\Delta}_{\epsilon}$. Moreover, standard adversarial training \eqref{eq:sat} minimizes $\max_{\delta\in\boldsymbol{\Delta}_{\epsilon}}\mathcal{L}(\boldsymbol{P},\boldsymbol{x}+\delta)$, which only guarantees robustness at isolated worst-case points (not neighborhoods). The region-based bound of ER-APT is strictly stronger when $\gamma+L\eta<\max_{\delta\in\boldsymbol{\Delta}_{\epsilon}}\mathcal{L}(\boldsymbol{P},\boldsymbol{x}+\delta)$. 
% \clearpage
\section{Experiments}
\subsection{Setups}

\begin{table*}[t]
  \centering
  
  % \setlength{\tabcolsep}{4pt}
  % \vspace{-3mm}
  \resizebox{\textwidth}{!}{
    \begin{tabular}{lcccccccccccc}
    \toprule
    \multicolumn{1}{l}{\multirow{2}[2]{*}{\textbf{Natural Acc (\%)}}} & \multicolumn{1}{c}{\multirow{2}[2]{*}{\textbf{ImageNet-1K}}} & \multicolumn{1}{c}{\multirow{2}[2]{*}{\textbf{Caltech101}}} & \multicolumn{1}{c}{\multirow{2}[2]{*}{\textbf{DTD}}} & \multicolumn{1}{c}{\multirow{2}[2]{*}{\textbf{EuroSAT}}} & \multicolumn{1}{c}{\multirow{2}[2]{*}{\textbf{OxfordPets}}} & \multicolumn{1}{c}{\multirow{2}[2]{*}{\textbf{FGVCAircraft}}} & \multicolumn{1}{c}{\multirow{2}[2]{*}{\textbf{Food101}}} & \multicolumn{1}{c}{\multirow{2}[2]{*}{\textbf{Flowers102}}} & \multicolumn{1}{c}{\multirow{2}[2]{*}{\textbf{StanfordCars}}} & \multicolumn{1}{c}{\multirow{2}[2]{*}{\textbf{SUN397}}} & \multicolumn{1}{c}{\multirow{2}[2]{*}{\textbf{UCF101}}} & \multicolumn{1}{c}{\multirow{2}[2]{*}{\textbf{Average}}} \\
    \multicolumn{1}{l}{} & \multicolumn{1}{c}{} & \multicolumn{1}{c}{} & \multicolumn{1}{c}{} & \multicolumn{1}{c}{} & \multicolumn{1}{c}{} & \multicolumn{1}{c}{} & \multicolumn{1}{c}{} & \multicolumn{1}{c}{} & \multicolumn{1}{c}{} & \multicolumn{1}{c}{} & \multicolumn{1}{c}{} & \multicolumn{1}{c}{} \\
    \midrule
    C-AVP~\cite{chen2023visual} & 46.27 & 90.40 & 29.20 & 18.13 & 56.40 & 1.33 & 1.07 & 56.17 & 14.83 & 54.70 & 0.97 & 33.59 \\
    APT~\cite{li2024one} & 52.63 & \textbf{92.93} & 54.50 & 33.40 & 83.70 & 14.77 & 62.50 & 86.63 & 51.90 & \textbf{65.67} & \textbf{69.40} &  60.73 \\
    AdvPT~\cite{zhang2024adversarial} &  24.53 & 68.70 & 43.77 &  53.33 &  46.27 & 10.07 & 18.47 & 56.03 &  14.87 & 33.13 & 36.60 & 36.89 \\
 AdvMaPLe~\cite{khattak2023maple} &  52.93 & 92.17 & 57.93 & 54.97 & 83.27 & 23.63 & 65.13 & \textbf{87.87} & \textbf{56.17} & 63.57 & 68.97  & 64.24\\
  AdvVLP~\cite{zhou2024fewshot}  &  53.23 & 92.37 & 57.53 & 15.50 & 82.93 & 23.27 & 43.30 & 87.70 & 56.00 & 63.90 & 69.10 & 58.62\\
 FAP~\cite{zhou2024fewshot}  & 52.53 & 91.10 & 55.17 & \textbf{81.70} & 81.90 & 23.50 & 64.03 & 86.27 & 54.23 & 62.37 & 65.70 & 65.32 \\
 
  \cellcolor{gray! 40} \bfseries ER-APT (ours) & \cellcolor{gray! 40} \textbf{54.07} & \cellcolor{gray! 40} 90.90 & \cellcolor{gray! 40} \textbf{57.93} & \cellcolor{gray! 40} 79.13 &  \cellcolor{gray! 40} \textbf{83.77} & \cellcolor{gray! 40} \textbf{25.03} & \cellcolor{gray! 40} \textbf{65.70} & \cellcolor{gray! 40} 86.83 & \cellcolor{gray! 40} 56.07 & \cellcolor{gray! 40} 63.93 & \cellcolor{gray! 40} 68.90 & \cellcolor{gray! 40} \textbf{66.60} \\
    \bottomrule
    \toprule
    \multicolumn{1}{l}{\multirow{2}[2]{*}{\textbf{PGD-100 Acc (\%)}}} & \multicolumn{1}{c}{\multirow{2}[2]{*}{\textbf{ImageNet-1K}}} & \multicolumn{1}{c}{\multirow{2}[2]{*}{\textbf{Caltech101}}} & \multicolumn{1}{c}{\multirow{2}[2]{*}{\textbf{DTD}}} & \multicolumn{1}{c}{\multirow{2}[2]{*}{\textbf{EuroSAT}}} & \multicolumn{1}{c}{\multirow{2}[2]{*}{\textbf{OxfordPets}}} & \multicolumn{1}{c}{\multirow{2}[2]{*}{\textbf{FGVCAircraft}}} & \multicolumn{1}{c}{\multirow{2}[2]{*}{\textbf{Food101}}} & \multicolumn{1}{c}{\multirow{2}[2]{*}{\textbf{Flowers102}}} & \multicolumn{1}{c}{\multirow{2}[2]{*}{\textbf{StanfordCars}}} & \multicolumn{1}{c}{\multirow{2}[2]{*}{\textbf{SUN397}}} & \multicolumn{1}{c}{\multirow{2}[2]{*}{\textbf{UCF101}}} & \multicolumn{1}{c}{\multirow{2}[2]{*}{\textbf{Average}}} \\
    \multicolumn{1}{l}{} & \multicolumn{1}{c}{} & \multicolumn{1}{c}{} & \multicolumn{1}{c}{} & \multicolumn{1}{c}{} & \multicolumn{1}{c}{} & \multicolumn{1}{c}{} & \multicolumn{1}{c}{} & \multicolumn{1}{c}{} & \multicolumn{1}{c}{} & \multicolumn{1}{c}{} & \multicolumn{1}{c}{} & \multicolumn{1}{c}{} \\
    \midrule
    
    C-AVP~\cite{chen2023visual} &  12.77 & 52.60 & 13.87 & 15.83 & 16.43 & 0.63 & 0.80 & 22.03 & 3.57 & 17.63 & 0.93 & 14.28\\
    APT~\cite{li2024one} & 2.07 & 30.23 & 10.47 &  0.87 & 4.40 & 1.27 & 2.63 & 8.97 & 1.60 & 3.67 & 4.40 & 6.42 \\
    AdvPT~\cite{zhang2024adversarial} & 1.47 & 9.63 & 5.70 &  0.17 &  0.23 &  0.43 & 0.73 & 0.80 & 0.33 & 2.37 & 0.53 & 2.04 \\
    AdvMaPLe~\cite{khattak2023maple} &  21.90 & 68.63 & 32.17 & 32.97 & 36.87 & 7.33 & 25.27 & 58.70 & 17.57 & 29.70 & 31.67  & 32.98 \\
   AdvVLP~\cite{zhou2024fewshot}  & 22.10 & 67.97 & \textbf{32.73} & 17.30 & 35.57 & 8.40 & 16.50 & 58.70 & 17.47 & 29.70 & 32.80  & 30.84\\
   
    FAP~\cite{zhou2024fewshot} & 22.90 & 67.33 & 31.33 & 39.73 & 41.00 & 7.97 & 26.67 & 61.47 & 19.23 & 30.27 & 32.80 & 34.61\\
    
   \cellcolor{gray! 40} \bfseries ER-APT (ours) & \cellcolor{gray! 40} \textbf{23.77} & \cellcolor{gray! 40} \textbf{70.53} & \cellcolor{gray! 40} 32.03 & \cellcolor{gray! 40} \textbf{42.50} & \cellcolor{gray! 40} \textbf{43.50} & \cellcolor{gray! 40} \textbf{8.53} & \cellcolor{gray! 40} \textbf{26.87} & \cellcolor{gray! 40} \textbf{64.43} & \cellcolor{gray! 40} \textbf{20.93} & \cellcolor{gray! 40} \textbf{30.70} & \cellcolor{gray! 40} \textbf{34.67} & \cellcolor{gray! 40} \textbf{36.22}\\
    \bottomrule
    \end{tabular}%
    } 
    \vspace{-3mm}
    \caption{\textbf{Accuracy (\%) of adversarial 16-shot learning on 11 datasets}. The table above presents the \textbf{Natural Accuracy}, while the table below reports the \textbf{Robust Accuracy} under PGD-100 attacks. Bolded numbers indicate the state-of-the-art results.}
  \label{tab:16shots}%
\end{table*}%

\textbf{Datasets.} To evaluate our method's effectiveness, we follow previous works \cite{zhou2024fewshot,khattak2023maple,zhou2022learning,zhou2022conditional} and utilize 11 diverse image recognition datasets covering multiple vision tasks. Specifically, the datasets comprise two generic-object datasets (ImageNet-1K~\cite{deng2009imagenet} and Caltech101~\cite{fei2004learning}),  a texture recognition dataset (DTD~\cite{cimpoi2014describing}), and five fine-grained object recognition datasets, including FGVCAircraft~\cite{maji2013fine}, OxfordPets~\cite{parkhi2012cats}, Flowers102~\cite{nilsback2008automated}, Food101~\cite{bossard2014food}, and StanfordCars~\cite{krause20133d}. Additionally, the collection includes datasets for scene recognition (SUN397~\cite{xiao2010sun}), action recognition (UCF101~\cite{soomro2012ucf101}), and satellite image classification (EuroSAT~\cite{helber2019eurosat}).

\noindent \textbf{Baselines.} To showcase the effectiveness of the proposed method, we compare it with Zero-shot CLIP~\cite{radford2021learning} capabilities as well as several existing adversarial prompt learning methods. First, we compare single-modal adversarial prompt learning methods. For visual prompts, we evaluate against C-AVP \cite{chen2023visual}. For textual prompts, we compare two methods: APT \cite{li2024one}, which focuses on learning robust text prompts instead of modifying model weights, and AdvPT \cite{zhang2024adversarial}, which first adopts the image encoder to generate AEs and then leverages learnable text prompts for alignment. Additionally, we compare various multimodal adversarial prompt learning methods. In this regard, we primarily follow the methods proposed by \citet{zhou2024fewshot}, which includes AdvVLP, AdvMaPLe~\cite{khattak2023maple} and FAP~\cite{zhou2024fewshot}.

\noindent \textbf{Implementation details.} Our experiments are conducted on the ViT-B/32 CLIP architecture, with results averaged over three random seeds. In cross-dataset evaluations, all models are trained for 5 epochs, while for other benchmark settings, training extends to 10 epochs. Optimization is performed using SGD with a momentum of 0.9, an initial learning rate of 0.0035, and a cosine learning rate scheduler with a warm-up phase during the first epoch.  For adversarial prompt learning, we employ token prompts of size 2 in both the vision and text branches across the first nine transformer blocks. More implementation details
are shown in the Appendix. Adversarial attacks are generated under the $\ell_{\infty}$ threat model using a 2-step PGD attack with a perturbation bound of $\epsilon = 1/255$ and a step size of $\alpha = 1/255$, following \cite{radford2021learning}. For our evolution-based region, the iteration count and step size match those of PGD to ensure fairness. The adversarial robustness is assessed using a 100-step PGD attack. For our proposed ER-APT method, we set the initial population size to $ \mathcal{N} = 9 $ and the mutation perturbation intensity to $ \phi = 0.1 $. For the dynamic loss weighting, the temperature parameter is set to $T = 1.0 $. The selection of these parameters and impact on ER-APT's performance are shown in the Appendix.

\subsection{Adversarial few-shot learning}
\label{sec:experiments_few_shot}
In this scenario, we assess the model’s capability to learn robust representations under extreme data scarcity. To achieve this, we fine-tune the model using only $\{1, 2, 4, 8, 16\}$ samples per class, ensuring minimal downstream data availability. In the main text, we primarily present the results under the 16-shot setting, as shown in Table~\ref{tab:16shots}, while the remaining experimental results are provided in the Appendix. From the results, it is evident that compared to previous methods, FAP~\cite{zhou2024fewshot} exhibits lower accuracy on nature examples in a few datasets. However, its robustness accuracy achieves state-of-the-art (SOTA) performance. In contrast, ER-APT further improves performance on natural examples compared to FAP, achieving the best accuracy across more datasets, with an average improvement of 1.28\% over FAP. It outperforms FAP in terms of robustness, attaining the best results on 10 out of 11 datasets, with an average gain of 1.61\%.

\begin{table*}[t]
  \centering

  \resizebox{\textwidth}{!}{
    \begin{tabular}{lcccccccccccc}
    \toprule
    \multicolumn{1}{l}{\multirow{2}[2]{*}{\textbf{Natural Acc (\%)}}} & \multicolumn{1}{c}{\multirow{2}[2]{*}{\textbf{ImageNet-1K}}} & \multicolumn{1}{c}{\multirow{2}[2]{*}{\textbf{Caltech101}}} & \multicolumn{1}{c}{\multirow{2}[2]{*}{\textbf{DTD}}} & \multicolumn{1}{c}{\multirow{2}[2]{*}{\textbf{EuroSAT}}} & \multicolumn{1}{c}{\multirow{2}[2]{*}{\textbf{OxfordPets}}} & \multicolumn{1}{c}{\multirow{2}[2]{*}{\textbf{FGVCAircraft}}} & \multicolumn{1}{c}{\multirow{2}[2]{*}{\textbf{Food101}}} & \multicolumn{1}{c}{\multirow{2}[2]{*}{\textbf{Flowers102}}} & \multicolumn{1}{c}{\multirow{2}[2]{*}{\textbf{StanfordCars}}} & \multicolumn{1}{c}{\multirow{2}[2]{*}{\textbf{SUN397}}} & \multicolumn{1}{c}{\multirow{2}[2]{*}{\textbf{UCF101}}} & \multicolumn{1}{c}{\multirow{2}[2]{*}{\textbf{Average}}} \\
    \multicolumn{1}{l}{} & \multicolumn{1}{c}{} & \multicolumn{1}{c}{} & \multicolumn{1}{c}{} & \multicolumn{1}{c}{} & \multicolumn{1}{c}{} & \multicolumn{1}{c}{} & \multicolumn{1}{c}{} & \multicolumn{1}{c}{} & \multicolumn{1}{c}{} & \multicolumn{1}{c}{} & \multicolumn{1}{c}{} & \multicolumn{1}{c}{} \\
    \midrule
    Zero-shot CLIP~\cite{radford2021learning} & \textbf{62.10 } & \textbf{91.50 } & \textbf{43.70 } & \textbf{45.20 } & \textbf{87.40 } & \textbf{19.20 } & \textbf{80.50 } & \textbf{66.90 } & \textbf{60.40 } & \textbf{62.10 } & \textbf{62.00 } & \multicolumn{1}{c}{\textbf{61.91 }} \\
    \midrule
    C-AVP~\cite{chen2023visual} & 44.87 & 85.47 & 30.23 & 25.17 & 74.20 & 7.13 & 56.53 & 43.17 & 27.27 & 41.97 & 44.60 & 43.69 \\
    APT~\cite{li2024one} & 12.23 & 53.57 & 11.93 & 11.33 & 13.97 & 7.83 & 7.30 & 13.73 & 5.90 & 14.73 & 18.30 & 15.53\\
    AdvPT~\cite{zhang2024adversarial} & 23.50 & 63.70 & 19.47 & 20.40 & 43.10 & 4.60 & 12.23 & 28.57 & 9.23 & 26.33 & 25.77 & 25.17 \\
AdvVLP~\cite{zhou2024fewshot} & 53.23 & 87.33 & 33.43 & 18.37 & 78.80 & 10.70 & 55.80 & 49.77 & 38.70 & 52.80 & 51.50 & 48.22 \\
  AdvMaPLe~\cite{khattak2023maple} & 52.93 & 88.23 & 30.87 & 17.60 & 77.87 & 11.10 & 56.67 & 52.90 & 36.70 & 52.53 & 50.97 & 48.03 \\
 FAP~\cite{zhou2024fewshot}  & 52.53 & 87.80 & 30.93 & 15.30 & 78.20 & 10.70 & 55.83 & 51.20 & 38.70 & 52.47 & 51.73 & 47.76 \\
   \cellcolor{gray! 40} \textbf{ER-APT (ours)} & \cellcolor{gray! 40} 53.67 & \cellcolor{gray! 40} 87.83 & \cellcolor{gray! 40} 34.80 & \cellcolor{gray! 40} 16.73 & \cellcolor{gray! 40} 79.83 & \cellcolor{gray! 40} 11.90 & \cellcolor{gray! 40} 56.57 & \cellcolor{gray! 40} 53.93 & \cellcolor{gray! 40} 41.80 & \cellcolor{gray! 40} 53.17 & \cellcolor{gray! 40} 53.10 & \cellcolor{gray! 40} 49.39 \\
    \bottomrule
    \toprule
    \multicolumn{1}{l}{\multirow{2}[2]{*}{\textbf{PGD-100 Acc (\%)}}} & \multicolumn{1}{c}{\multirow{2}[2]{*}{\textbf{ImageNet-1K}}} & \multicolumn{1}{c}{\multirow{2}[2]{*}{\textbf{Caltech101}}} & \multicolumn{1}{c}{\multirow{2}[2]{*}{\textbf{DTD}}} & \multicolumn{1}{c}{\multirow{2}[2]{*}{\textbf{EuroSAT}}} & \multicolumn{1}{c}{\multirow{2}[2]{*}{\textbf{OxfordPets}}} & \multicolumn{1}{c}{\multirow{2}[2]{*}{\textbf{FGVCAircraft}}} & \multicolumn{1}{c}{\multirow{2}[2]{*}{\textbf{Food101}}} & \multicolumn{1}{c}{\multirow{2}[2]{*}{\textbf{Flowers102}}} & \multicolumn{1}{c}{\multirow{2}[2]{*}{\textbf{StanfordCars}}} & \multicolumn{1}{c}{\multirow{2}[2]{*}{\textbf{SUN397}}} & \multicolumn{1}{c}{\multirow{2}[2]{*}{\textbf{UCF101}}} & \multicolumn{1}{c}{\multirow{2}[2]{*}{\textbf{Average}}} \\
    \multicolumn{1}{l}{} & \multicolumn{1}{c}{} & \multicolumn{1}{c}{} & \multicolumn{1}{c}{} & \multicolumn{1}{c}{} & \multicolumn{1}{c}{} & \multicolumn{1}{c}{} & \multicolumn{1}{c}{} & \multicolumn{1}{c}{} & \multicolumn{1}{c}{} & \multicolumn{1}{c}{} & \multicolumn{1}{c}{} & \multicolumn{1}{c}{} \\
    \midrule
    Zero-shot CLIP~\cite{radford2021learning} & 1.57 & 26.23 & 5.07 & 0.03 & 3.27 & 0.00 & 5.03 & 1.73 & 0.30 & 1.20 & 2.47 & 4.26 \\
    \midrule
    C-AVP~\cite{chen2023visual} & 11.67& 48.07 & 12.93 & 4.57 & 19.03 & 0.83 & 9.70 & 16.20 & 2.90 & 12.77 & 10.47 & 13.56 \\
    APT~\cite{li2024one} & 0.90 & 7.70 & 3.47 & 0.00 & 1.10 & 0.63 & 0.10 & 0.67 & 0.00 & 2.37 & 0.33 & 1.57\\
    AdvPT~\cite{zhang2024adversarial} & 0.33 & 3.47 & 3.30 & 0.17 & 0.87 & 0.00 &  0.00 & 0.60 & 0.43 & 0.40 & 0.27 & 0.89 \\
    AdvVLP~\cite{zhou2024fewshot} & 22.10 & 62.97 & 18.60 & \textbf{10.67} & 40.83 & 2.73 & 17.83 & 25.23 & 10.97 & 21.67 & 22.10 & 23.25 \\
   AdvMaPLe~\cite{khattak2023maple} & 21.90 & 64.90 & 17.50 & 10.53 & 42.83 & 2.73 & 18.53 & 28.73 & 10.43 & 21.90 & 23.20 & 23.93\\
   
   FAP~\cite{zhou2024fewshot}  & 22.90 & 65.43 & 16.93 & 9.97 & 43.77 & 2.77 & 19.60 & 27.23 & 11.80 & 22.40 & 23.77 & 24.23 \\
   \cellcolor{gray! 40} \textbf{ER-APT (ours)} & \cellcolor{gray! 40} \textbf{23.20} & \cellcolor{gray! 40} \textbf{68.03} & \cellcolor{gray! 40} \textbf{19.80} & \cellcolor{gray! 40} 10.47 & \cellcolor{gray! 40} \textbf{45.93} & \cellcolor{gray! 40} \textbf{3.80} & \cellcolor{gray! 40} \textbf{19.77} & \cellcolor{gray! 40} \textbf{30.03} & \cellcolor{gray! 40} \textbf{13.07} & \cellcolor{gray! 40} \textbf{22.60} & \cellcolor{gray! 40} \textbf{24.27} & \cellcolor{gray! 40} \textbf{25.54} \\
    \bottomrule
    \end{tabular}%
    }
    \vspace{-3mm}
    \caption{Cross-dataset generalization from ImageNet-1K to various downstream recognition datasets. We report the mean and standard deviation of natural and robust (PGD-100) accuracy. Bolded numbers denote the state-of-the-art results.}
  \label{tab:cross_dataset}%
\end{table*}%
\begin{table}[t]
  \centering
  \vspace{-10pt}

    \resizebox{\linewidth}{!}{
    \begin{tabular}{lcccc}
    \toprule
    \multirow{2}[4]{*}{\textbf{Method}} & \multicolumn{2}{c}{\textbf{Base Class}} & \multicolumn{2}{c}{\textbf{New Class}} \\
\cmidrule{2-5}          & Base Nat Acc & Base Adv Acc & New Nat Acc & New Adv Acc \\
    \midrule
    C-AVP~\cite{chen2023visual} & 31.68 & 14.43 & 30.39 & 13.36 \\
    APT~\cite{li2024one} & 18.21  & 3.80 & 13.99 & 3.07 \\
    AdvPT~\cite{zhang2024adversarial} & 43.87 & 3.50 & 44.94 & 8.84 \\
    AdvVLP~\cite{zhou2024fewshot} & 58.95 & 32.37 & 46.92 & 21.61 \\
    AdvMaPLe~\cite{khattak2023maple} & 60.38 & 30.69 & 46.18 & 20.25 \\
     FAP~\cite{zhou2024fewshot} & 70.52 & 38.05 & 49.58 & 21.86 \\
    \cellcolor{gray! 40} \textbf{ER-APT (ours)} & \cellcolor{gray! 40} 
 \textbf{71.76} & \cellcolor{gray! 40}  \textbf{41.81} & \cellcolor{gray! 40} \textbf{53.88} & \cellcolor{gray! 40} 
 \textbf{26.29}\\
    \bottomrule
    \end{tabular}%
    }
    \caption{\textbf{Adversarial base-to-new Generalization performance.} We report the average result of the Base Natural Accuracy (\%), Base Adversarial Accuracy (\%), New Natural Accuracy (\%), and New Adversarial Accuracy (\%) on 11 datasets. Detailed results for each dataset are provided in the Appendix.}
  \label{tab:base2new}
\end{table}%

\subsection{Adversarial base-to-new generalization}
\label{sec:experiments_base2new}
We follow FAP and adopt a more challenging adversarial base-to-new generalization setting, where datasets are divided into base and new subclasses. In this setting, models are trained on a 16-shot dataset from the base classes and then evaluated on both base and new classes. Since the number of categories in most datasets is significantly smaller than the number of samples per class, models must extract intrinsic features from each dataset and learn robust representations from limited examples to effectively generalize to a large volume of test data. From Table~\ref{tab:base2new}, We observe that our proposed ER-APT outperforms the state-of-the-art method FAP not only in both Base Class and New Class, but also in terms of Natural Accuracy and Robust Accuracy. Specifically, on the Base Class, our method surpasses FAP by 1.24\% and 3.76\%, respectively. When evaluating the model on New Classes after conducting adversarial prompt learning on the Base Class, our approach achieves significantly higher performance than existing methods, exceeding them by 4.3\% and 4.43\%, respectively. 
% These results strongly demonstrate the effectiveness of our method in leveraging Evolution-Region to generate diverse perturbations for adversarial prompt learning, thereby enhancing both model robustness and generalization.

\subsection{Adversarial Cross-Dataset Evaluation}
\label{sec:experiments_cross_dataset}
Beyond comparisons with baselines under few-shot settings, we also evaluate our method in zero-shot scenarios, where robustness is assessed through cross-dataset evaluations. We adopt a 16-shot setting, where the CLIP model first undergoes adversarial prompt tuning on ImageNet-1K. Subsequently, the tuned model is evaluated for cross-dataset generalization across 10 downstream datasets. The detailed results are presented in Table~\ref{tab:cross_dataset}. It is evident that the zero-shot CLIP~\cite{radford2021learning} model achieves the highest accuracy on natural examples. However, it exhibits almost no defense against AEs. In contrast, models undergoing adversarial prompt tuning experience a slight decline in accuracy on natural examples but gain significantly improved robustness. This process seeks to strike a balance between clean accuracy and robustness. Among all adversarial prompt tuning methods, our approach achieves the best performance in both natural accuracy and robust accuracy on downstream datasets, surpassing the previous state-of-the-art method by 1.17\% and 1.31\%, respectively. This further demonstrates the effectiveness of our evolution-based method in enhancing generalization.

\begin{table*}[t]
  \centering
 
    \resizebox{\linewidth}{!}{
    \begin{tabular}{lcccccccccccc}
    \toprule
    \multicolumn{1}{c}{\multirow{2}[2]{*}{\textbf{AutoAttack}}} & \multirow{2}[2]{*}{\textbf{ImageNet-1K}} & \multirow{2}[2]{*}{\textbf{Caltech101}} & \multirow{2}[2]{*}{\textbf{DTD}} & \multirow{2}[2]{*}{\textbf{EuroSAT}} & \multirow{2}[2]{*}{\textbf{OxfordPets}} & \multirow{2}[2]{*}{\textbf{FGVCAircraft}} & \multirow{2}[2]{*}{\textbf{Food101}} & \multirow{2}[2]{*}{\textbf{Flowers102}} & \multirow{2}[2]{*}{\textbf{StanfordCars}} & \multirow{2}[2]{*}{\textbf{SUN397}} & \multirow{2}[2]{*}{\textbf{UCF101}} & \multirow{2}[2]{*}{\textbf{Average}} \\
          &       &       &       &       &       &       &       &       &       &       &       &  \\
    \midrule
     FAP  & 41.87 & 24.53 & 
     33.33 & 42.30 & 27.40 & 16.27 & 52.63 & 48.17 & 37.60 & \textbf{49.83} & 38.10 & 37.45 \\
    \cellcolor{gray! 40} \bfseries ER-APT (ours) & \cellcolor{gray! 40} \textbf{42.23} & \cellcolor{gray! 40} \textbf{27.07} & \cellcolor{gray! 40} \textbf{33.37} & \cellcolor{gray! 40} \textbf{49.10} & \cellcolor{gray! 40} \textbf{33.93} & \cellcolor{gray! 40} \textbf{19.30} & \cellcolor{gray! 40} \textbf{53.57} & \cellcolor{gray! 40} \textbf{49.63} & \cellcolor{gray! 40} \textbf{38.63} & \cellcolor{gray! 40} 49.30 & \cellcolor{gray! 40} \textbf{52.03} & \cellcolor{gray! 40} \textbf{40.74} \\
    \bottomrule
    \end{tabular}%
    }
     \caption{\textbf{Zero-shot adversarial robustness (\%) with Auto-Attack adversarial perturbation.} We perform adversarial prompt tuning using perturbations generated by PGD attack and evaluate the model under AutoAttack perturbations with $\epsilon=1/255$.}

  \label{tab:auto_attack}%
\end{table*}%

\label{sec:experiments_autuattack}
\subsection{Zero-shot Evaluation under Auto-Attack}
We employ the more powerful Auto-Attack~\cite{croce2020reliable} to evaluate our adapted model. We follow the adversarial few-shot prompt learning setup with a 16-shot setting, utilizing adversarial perturbations generated by the PGD attack. After tuning the model, we evaluate its performance against perturbations generated by AutoAttack. The AutoAttack used in our experiments is the standard version, which consists of an ensemble of APGD-CE, APGD-DLR, FAB~\cite{croce2020minimally}, and Square Attack~\cite{andriushchenko2020square}. Moreover, AutoAttack provides a strong and reproducible benchmark for evaluating adversarial robustness, ensuring consistency across different models and experiments. Its ability to systematically explore worst-case perturbations makes it a reliable tool for assessing the true robustness of a model. Notably, AutoAttack utilizes a fractional attack generator that adaptively adjusts the step size $\alpha$, enhancing its effectiveness for zero-shot robustness evaluation. As shown in Table~\ref{tab:auto_attack}, we observe that ER-APT surpasses FAP in robustness against AutoAttack on 10 out of 11 datasets, achieving an average improvement of 3.29\%. This demonstrates that even under the stronger AutoAttack, our Evolution-Region method remains SOTA in robustness.

% This strongly demonstrates that when using the more powerful AutoAttack for a more realistic evaluation of model robustness, our Evolution-Region-based method still achieves SOTA performance.

\subsection{Ablation Study}
\begin{figure}[t]
    \centering
    \includegraphics[width=1\linewidth]{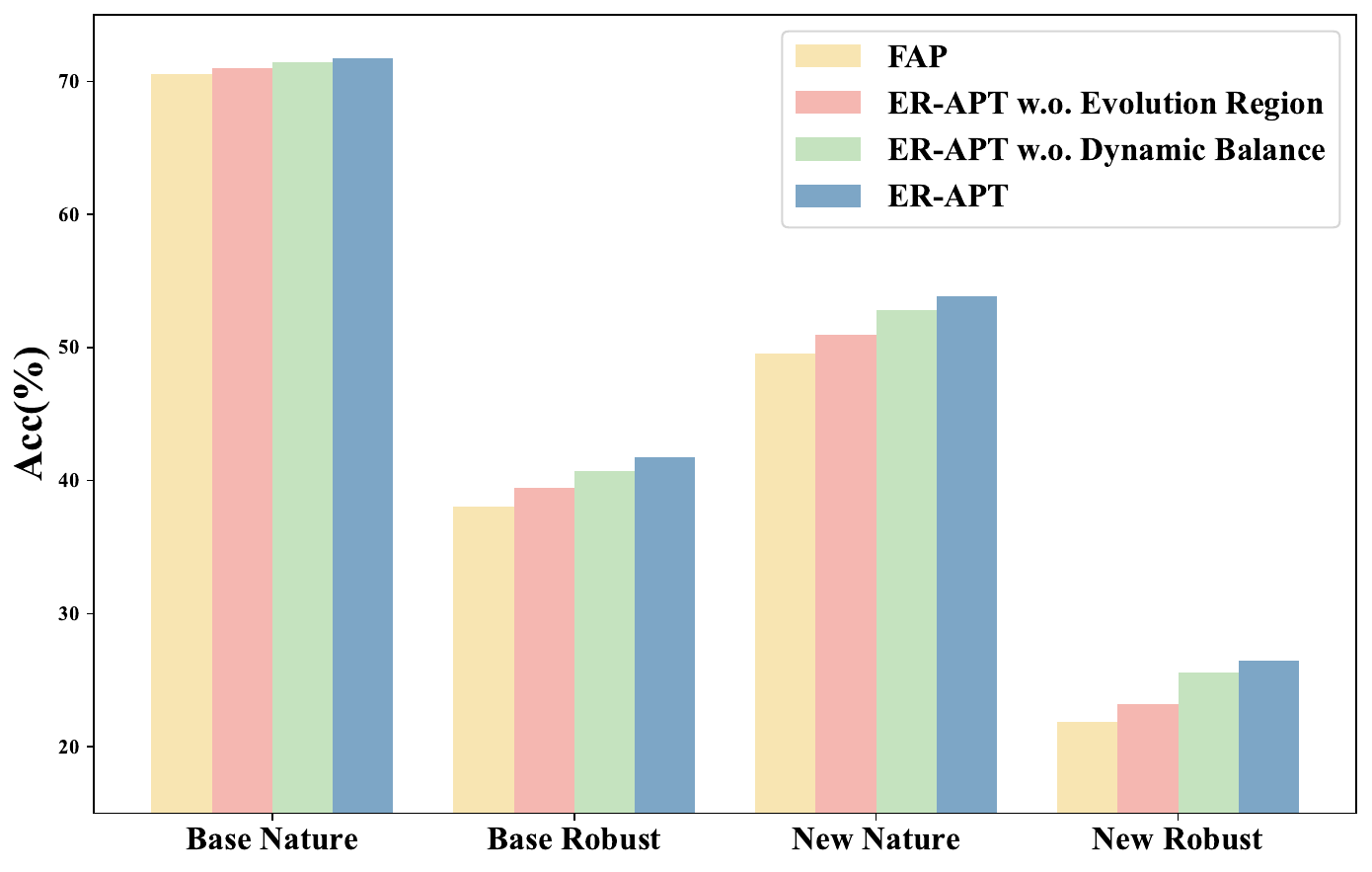}
    \caption{\textbf{Ablation Study.} The two innovations, \textbf{Evolution-based Region} and \textbf{Dynamic Loss Weighting}, are each ablated to compare their adversarial base-to-new generalization.}
    \label{fig:ablation}
\end{figure}
Our proposed ER-APT proposes two key advancements over previous adversarial prompt tuning methods: Evolution-based Region and Dynamic Loss Weighting. To evaluate the contribution of each component to the overall effectiveness of our method, we conduct comprehensive ablation studies on adversarial base-to-new generalization. Specifically, we systematically remove each component from ER-APT and analyze its impact on multiple performance metrics, including Base and New Class accuracy, Natural Accuracy, and Robust Accuracy. The detailed results, presented in Fig.~\ref{fig:ablation}, clearly demonstrate that eliminating either of the two improvements leads to a decline across almost all evaluation metrics, underscoring their critical role in enhancing ER-APT’s performance. Notably, the performance degradation is especially severe when the Evolution-based Region component is removed, highlighting its paramount importance in strengthening both the robustness and generalization capabilities of our method. 
% These findings affirm that both enhancements work synergistically to improve adversarial prompt tuning, with the Evolution-based Region playing a particularly significant role.

% Our proposed ER-APT introduces two key improvements over previous adversarial prompt tuning methods: \textbf{Evolution-based Region} and \textbf{dynamic loss weighting}. To assess the impact of each improvement on the effectiveness of our approach, we conduct ablation studies on adversarial base-to-new generalization. Specifically, we systematically remove each component from ER-APT and compare the changes in Base and New Classes as well as Natural Accuracy and Robust Accuracy. The detailed results are presented in Fig.~\ref{fig:ablation}. It is evident that eliminating either of the two improvements leads to a decline across almost all evaluation metrics, underscoring their critical role in enhancing ER-APT’s performance. Notably, the performance degradation is especially severe when the Evolution-based Region component is removed, highlighting its paramount importance in strengthening both the robustness and generalization capabilities of our method. 

\subsection{Computation Cost}
\begin{table}[t]
  \centering

    \resizebox{0.95\linewidth}{!}{
    \begin{tabular}{lcccccccccccc}
    \toprule
    \multicolumn{1}{c}{\multirow{2}[2]{*}{\textbf{Training Time(mins)}}} & \multirow{2}[2]{*}{\textbf{Caltech101}} & \multirow{2}[2]{*}{\textbf{DTD}} & \multirow{2}[2]{*}{\textbf{OxfordPets}} & \multirow{2}[2]{*}{\textbf{UCF101}} \\
          &       &       &       &   \\
    \midrule
     FAP & 25 & 21 & 19 & 35\\
    \cellcolor{gray! 40} \bfseries ER-APT (ours) & \cellcolor{gray! 40} 42 & \cellcolor{gray! 40} 30 & \cellcolor{gray! 40} 27 & \cellcolor{gray! 40} 55\\
    \bottomrule
    \end{tabular}%
    }
    \caption{{Comparison of training time between FAP and ER-APT, evaluated on the selected datasets: Caltech101, DTD, OxfordPets, and UCF101.}}
  \label{tab:time_cost}%
  \vspace{-3mm}
\end{table}%

In addition to comparing robustness and generalization with existing methods, we further evaluate the training time efficiency of our approach compared to the state-of-the-art method FAP~\cite{zhou2024fewshot}. We maintain the 16-shot adversarial prompt tuning setup and conduct experiments on the selected datasets, Caltech101~\cite{fei2004learning}, DTD~\cite{cimpoi2014describing}, OxfordPets~\cite{parkhi2012cats} and UCF101~\cite{soomro2012ucf101}. The detailed training time comparison is presented in Table~\ref{tab:time_cost}, where it is evident that ER-APT requires \textbf{1.42} to \textbf{1.68} times the training time of FAP. This additional time primarily stems from the evolution-based region, which generates more diverse AEs. 
Although our ER-APT incurs a slightly higher training cost than FAP, our ER-APT consistently achieves state-of-the-art robustness and generalization across a wide range of tasks and datasets, making the trade-off highly favorable.
% , it consistently achieves state-of-the-art robustness and generalization across diverse tasks and datasets. 
% Notably, it demonstrates superior performance in adversarial few-shot learning (Sec~\ref{sec:experiments_few_shot}), adversarial base-to-new generalization (Sec~\ref{sec:experiments_base2new}), adversarial cross-dataset generalization (Sec~\ref{sec:experiments_cross_dataset}), and zero-shot adversarial evaluation under AutoAttack (Sec~\ref{sec:experiments_autuattack}).

\section{Conclusion}
In this work, we propose an evolution-based region adversarial prompt tuning method, called ER-APT, to address the limitations of existing adversarial prompt tuning methods for large vision-language models (VLMs).
Unlike previous methods that rely on single-gradient direction perturbations, ER-APT integrates gradient-based attacks with genetic evolution, enhancing the diversity and effectiveness of adversarial examples. By leveraging selection, mutation, and crossover mechanisms, our approach optimizes adversarial examples to cover a broader perturbation space, leading to more robust prompt tuning. Additionally, the dynamic loss weighting strategy further improves the balance between robustness and accuracy during training, improving the adversarial robustness.
Extensive experiments conducted on multiple benchmark datasets confirm that ER-APT consistently outperforms state-of-the-art adversarial prompt tuning methods, achieving superior adversarial robustness while preserving strong generalization capabilities. Our results show that ER-APT enhances resilience to adversarial perturbations while ensuring robust transferability across diverse tasks and data distributions.
 Our findings highlight the importance of diverse and adaptive FAT for improving the robustness. 
 % paving the way for more resilient vision-language models in real-world applications.
% \input{sec/2_formatting}
% \input{sec/3_finalcopy}
\clearpage
{
    \small
    \bibliographystyle{ieeenat_fullname}
    \bibliography{main}
}

\clearpage
\appendix

\section{Additional Implementation Details}
\subsection{Environment and Device}
Our experiments are conducted in an environment with PyTorch 1.10.1, CUDA 11.3, and Python 3.8. The computations are performed on either an NVIDIA A100-SXM4-40GB or an NVIDIA A100-SXM4-80GB GPU.

\subsection{Hand-crafted Prompt Templates} 
\label{Hand-crafted Prompts templates}
We report the hand-crafted prompt templates used in Zero-shot CLIP and our method for initialization on 11 image recognition datasets in Table~\ref{tab:static_template}. 

\begin{table}[ht]
  \centering
  \resizebox{\linewidth}{!}{
    \begin{tabular}{cc}
    \toprule
    Dataset & \multicolumn{1}{c}{Template} \\
    \midrule
    ImageNet-1K & \texttt{"a photo of a \{\}."} \\
    Caltech101 & \texttt{"a photo of a \{\}."} \\
    DTD   & \texttt{"\{\} texture."} \\
    EuroSAT & \texttt{"a centered satellite photo of \{\}."} \\
    OxfordPets & \texttt{"a photo of a \{\}, a type of pet."} \\
    FGVCAircraft & \texttt{"a photo of a \{\}, a type of aircraft."} \\
    Food101 & \texttt{"a photo of a \{\}, a type of food."} \\
    Flowers102 & \texttt{"a photo of a \{\}, a type of flower."} \\
    StanfordCars & \texttt{"a photo of a \{\}."} \\
    SUN397 & \texttt{"a photo of a \{\}."} \\
    UCF101 & \texttt{"a photo of a person doing \{\}."} \\
    \bottomrule
    \end{tabular}%
    }
    \vspace{-3mm}
  \caption{Hand-crafted text template for static text supervision of different datasets.}
  \label{tab:static_template}%
  \vspace{-4mm}
\end{table}%

\subsection{Comparison between APT and AdvPT}
APT~\cite{li2024one} and AdvPT~\cite{zhang2024adversarial} are both adversarial prompt tuning methods based on the CoOp~\cite{zhou2022conditional} framework, which updates only the learnable tokens in the category-specific prompt without modifying the model parameters. The key difference between the two lies in the generation of adversarial examples during training. In APT, adversarial examples are dynamically updated at each iteration as the text prompt evolves. In contrast, AdvPT generates all adversarial examples in advance using the zero-shot CLIP model before training begins. Regarding the length of text prompts, we typically follow the practice of using 16 context tokens, with an additional class token appended at the end of the context.
\section{Hyperparameters}

\begin{table}[t]
\begin{center}
\small
\renewcommand\arraystretch{1}
\setlength{\tabcolsep}{4pt}
    \resizebox{\linewidth}{!}{
		\begin{tabular}{ @{\extracolsep{\fill}} c|cc|cc|cc} 
        \toprule[0.3mm]
			&  \multicolumn{2}{c}{\textbf{OxfordPets~\cite{parkhi2012cats}}} & \multicolumn{2}{c}{\textbf{Caltech101~\cite{fei2004learning})}} & \multicolumn{2}{c}{DTD~\cite{cimpoi2014describing}} \\
			\cmidrule{2-7} 
			\multirow{-2}{*}{\textbf{$\mathcal{N}$}} & {Nature Acc} & {Robust Acc} & {Nature Acc} & {Robust Acc} & {Nature Acc} & {Robust Acc} \\
			\midrule
            6 & 83.33 & 43.13 & \textbf{91.10} & 69.07 & 57.17 & 31.43 \\
            \cellcolor{gray! 40} 9 & \cellcolor{gray! 40} \textbf{83.77} & \cellcolor{gray! 40} \textbf{43.50} & \cellcolor{gray! 40} 90.90 & \cellcolor{gray! 40} 70.53 & \cellcolor{gray! 40} \textbf{57.93} & \cellcolor{gray! 40} 32.03 \\
            12 & 83.70 & 42.57 & 90.67 & \textbf{70.97} & 57.87 & 31.97 \\
            15 & 83.43 & 43.20 & 90.80 & 70.63 & 57.90 & \textbf{32.30} \\
   \bottomrule[0.3mm]
\end{tabular}}
\end{center}
\vspace{-6mm}
\caption{\textbf{The impact of population size $\mathcal{N}$ in ER-APT initialization on its performance.} The gray color represents the selected hyperparameter.}
\vspace{-3mm}
\label{tab:appendix_population_size}
\end{table}
\begin{table}[t]
\begin{center}
\small
\renewcommand\arraystretch{1}
\setlength{\tabcolsep}{4pt}
    \resizebox{\linewidth}{!}{
		\begin{tabular}{ @{\extracolsep{\fill}} c|cc|cc|cc} 
        \toprule[0.3mm]
			&  \multicolumn{2}{c}{\textbf{OxfordPets~\cite{parkhi2012cats}}} & \multicolumn{2}{c}{\textbf{Caltech101~\cite{fei2004learning})}} & \multicolumn{2}{c}{DTD~\cite{cimpoi2014describing}} \\
			\cmidrule{2-7} 
			\multirow{-2}{*}{\textbf{$\phi$}} & {Nature Acc} & {Robust Acc} & {Nature Acc} & {Robust Acc} & {Nature Acc} & {Robust Acc} \\
			\midrule
            0.05 & 83.43 & 41.30 & 91.20 & 69.97 & 57.37 & \textbf{32.43}\\
            \cellcolor{gray! 40} 0.1 & \cellcolor{gray! 40} \textbf{83.77} & \cellcolor{gray! 40} \textbf{43.50} & \cellcolor{gray! 40} 90.90 & \cellcolor{gray! 40} \textbf{70.53} & \cellcolor{gray! 40} \textbf{57.93} & \cellcolor{gray! 40} 32.03 \\
            0.2 & 83.2 & 43.2 & \textbf{91.27} & 69.93 & 57.70 & 31.70 \\
   \bottomrule[0.3mm]
\end{tabular}}
\vspace{-3mm}
\caption{\textbf{The impact of the population mutation perturbation intensity parameter $\phi$ in ER-APT on its performance.} The gray color indicates the selected hyperparameter.}
\label{tab:mutation_params}
\vspace{-6mm}
\end{center}
\end{table}
\begin{table}[t]
\begin{center}
\small
\renewcommand\arraystretch{1}
\setlength{\tabcolsep}{4pt}
    \resizebox{\linewidth}{!}{
		\begin{tabular}{ @{\extracolsep{\fill}} c|cc|cc|cc} 
        \toprule[0.3mm]
			&  \multicolumn{2}{c}{\textbf{OxfordPets~\cite{parkhi2012cats}}} & \multicolumn{2}{c}{\textbf{Caltech101~\cite{fei2004learning})}} & \multicolumn{2}{c}{DTD~\cite{cimpoi2014describing}} \\
			\cmidrule{2-7} 
			\multirow{-2}{*}{\textbf{$T$}} & {Nature Acc} & {Robust Acc} & {Nature Acc} & {Robust Acc} & {Nature Acc} & {Robust Acc} \\
			\midrule
            0.5 & 83.70 & 42.27 & \textbf{91.13} & 70.03 & 55.17 & 30.63 \\
            \cellcolor{gray! 40} 1.0 & \cellcolor{gray! 40} \textbf{83.77} & \cellcolor{gray! 40} \textbf{43.50} & \cellcolor{gray! 40} 90.90 & \cellcolor{gray! 40} 70.53 & \cellcolor{gray! 40} \textbf{57.93} & \cellcolor{gray! 40} \textbf{32.03} \\
            2.0 & 83.50 & 43.07 & 90.80 & \textbf{71.57} & 57.57 & 31.37 \\
   \bottomrule[0.3mm]
\end{tabular}}
\vspace{-3mm}
\end{center}
\caption{\textbf{The impact of the temperature parameter $T$ in dynamic loss weighting on ER-APT performance.} The gray color indicates the selected hyperparameter.}
\label{tab:temperature}
\end{table}

\textbf{Initialization of evolutionary population size $\mathcal{N}$.} The population size $\mathcal{N}$ controls the diversity of adversarial examples and influences computational complexity. In general, increasing $\mathcal{N}$ allows a broader exploration of the attack space, leading to enhanced robustness. As shown in \cref{tab:appendix_population_size}, when $\mathcal{N} = 9$, the method achieves strong robustness (OxfordPets:43.50\%, Caltech101: 70.53\%), and further increasing $\mathcal{N}$ to 15 provides only marginal improvement (Caltech101: 70.63\%, DTD: 32.30\%). However, a larger population size significantly increases computational cost, making it impractical in certain scenarios. Given that $\mathcal{N} = 9$ already achieves competitive robustness while maintaining computational efficiency, we select it as the optimal population size.

\textbf{Mutation perturbation intensity $\phi$.} The population mutation intensity $\phi$ determines the diversity of adversarial examples during the evolutionary process. As shown in \cref{tab:mutation_params}, selecting $\phi = 0.1$ achieves high robustness across datasets, particularly for Caltech101 (Robust Acc: 70.53\%) and OxfordPets (Robust Acc: 43.50\%). When $\phi = 0.05$, the mutation strength is insufficient, leading to a limited search space and lower robustness (Caltech101: 69.97\%, OxfordPets: 41.30\%). On the other hand, increasing $\phi$ to 0.2 introduces excessive perturbation, which may cause instability in the optimization process, resulting in a slight decline in robustness (Caltech101: 69.97\%). Given these observations, $\phi = 0.1$ is selected as a balanced choice that ensures sufficient perturbation diversity while maintaining stable training.

\textbf{Dynamic loss weighting temperature $T$.} The temperature parameter $T$ in dynamic loss weighting determines how the balance between clean accuracy and adversarial robustness is adjusted. As shown in \cref{tab:temperature}, setting $T = 1.0$ yields stable performance across datasets, particularly for OxfordPets (Robust Acc: 43.50\%) and DTD (Robust Acc: 32.03\%). When $T = 0.5$, the loss weight changes more abruptly, which may destabilize training and result in lower robustness (Caltech101: 70.03\%, DTD: 30.63\%). Conversely, increasing $T$ to 2.0 results in smoother loss weight adjustments, but it also weakens the dynamic weighting of the loss, ultimately leading to performance degradation (DTD: 31.37\%). Given these observations, $T = 1.0$ is selected to ensure a stable and effective balance between accuracy and robustness.

\section{Training Pipeline for ER-APT}
Based on the detailed methodology presented in the main text, we further provide a more intuitive training pipeline for Evolution-based Region Adversarial Prompt Learning, as outlined in \cref{alg:er-apt}.

\section{Detailed experimental results for each dataset.}
Due to space limitations in the main text, we present only the 16-shot results for adversarial few-shot learning. However, in \cref{tab:few_shot_natural_accuracy_appendix} and \cref{tab:few_shot_robust_accuracy_appendix}, we provide a detailed breakdown of the natural accuracy and robustness accuracy across different shot settings (1, 2, 4, 8, and 16) for each dataset. Additionally, for the adversarial Base-to-New generalization experiments, we present only the averaged results in the main text. However, in \cref{tab:base2new_appendix}, we provide detailed results for each dataset.

\newpage

% {
%     \small
%     \bibliographystyle{ieeenat_fullname}
%     \bibliography{main}
% }

\begin{algorithm*}[t]
\caption{Evolution-based Region Adversarial Prompt Tuning (ER-APT)}
\label{alg:er-apt}
\textbf{Input:} Few-shot dataset $\mathcal{S}$, CLIP model parameters $\theta = \{\theta_{\mathcal{I}}, \theta_{\mathcal{T}}\}$, prompt vectors $\boldsymbol{P} = \{\boldsymbol{P}_v, \boldsymbol{P}_t\}$,  text description $\boldsymbol{t}$, population size $\mathcal{N}$, mutation perturbation intensity $\phi$, learning rate $\alpha$, dynamic loss weighting temperature $T$, adversarial budget $\epsilon$, the initial loss balancing parameters $\alpha_{init},\beta_{init}$.\\
\textbf{Output:} Optimized prompt vectors $\boldsymbol{P}^*$.
\begin{algorithmic}[1]
\FOR{each training epoch}
    \FOR{each minibatch $(\boldsymbol{x}, t) \sim \mathcal{S}$}
        \STATE \textit{\# Calculate image and word embeddings}
        \STATE $\boldsymbol{h}(\boldsymbol{x}, \boldsymbol{P}_v)=\left\{c_{\mathrm{cls}}, h_1(\mathbf{x}), \ldots, h_K(\mathbf{x}), \boldsymbol{P}_v\right\}$
        \STATE $\boldsymbol{d}\left(t_j, \boldsymbol{P}_t \right)=\left\{d_1\left(t_j\right), \ldots, d_M\left(t_j\right), \boldsymbol{P}_t \right\}$ 
        \STATE \textit{\# Generate clean visual and text representations}
        \STATE $\boldsymbol{z}_{(v,\boldsymbol{P}_v)}=f_{I}\left(\boldsymbol{h}(\boldsymbol{x}, \boldsymbol{P}_v) ; \mathcal{W}_{v}\right)$
        \STATE $\boldsymbol{z}_t^{(j,\boldsymbol{P}_t)}=f_{t}\left(\boldsymbol{d}\left(t_j, \boldsymbol{P}_t\right) ; \mathcal{W}_{t}\right)$
        \STATE \textit{\# Initialize adversarial population}
        \STATE $\mathcal{X}^0 \gets \{\boldsymbol{x} + \boldsymbol{\delta}_i^0\}_{i=1}^{\mathcal{N}}, \quad \boldsymbol{\delta}_i^0 \sim \mathcal{U}(-\epsilon, \epsilon)$
        
        \FOR{each attack iteration $t$}
            \STATE \textit{\# Optimize adversarial examples using gradients}
            \STATE $\boldsymbol{\delta}_i^{t}=\Pi_{[-\epsilon, \epsilon]} [\boldsymbol{\delta}_i^{t-1}+\alpha \operatorname{sign}(\nabla_{\boldsymbol{\delta}_i^{t-1}} \mathcal{L}(f_{\boldsymbol{P}}(\boldsymbol{x}+\boldsymbol{\delta}_i^{t-1}), y))]$
            \STATE \textit{\# Fitness evaluation and selection}
            \STATE Compute fitness scores $\mathcal{F}(\boldsymbol{x}+\boldsymbol{\delta}_i^{t}) = \mathcal{L} \left( f_{\boldsymbol{P}} (\boldsymbol{x}+\boldsymbol{\delta}_i^{t}), y \right).$
            \STATE Select top $\mathcal{N}/3$ samples: $\mathcal{X}^{\text{selected}, t}$
            
            \STATE \textit{\# Mutation}
            \FOR{each selected sample $\boldsymbol{x}_{s_i}$}
                \STATE $\boldsymbol{\delta}^t_{m_i} = \Pi_{[-\epsilon, \epsilon]}[\boldsymbol{\delta}^t_{s_i} + \xi], \quad \xi \sim \mathcal{U}(-\phi \times \epsilon, \phi \times \epsilon)$
            \ENDFOR
            \STATE $\mathcal{X}^{\text{mutated}, t} \gets \{\boldsymbol{x} + \boldsymbol{\delta}_{m_i}^t\}_{i=1}^{\mathcal{N}/3}$
            
            \STATE \textit{\# Crossover}
            \FOR{$i=1$ to $\mathcal{N}/3$}
                \STATE Randomly sample $\boldsymbol{\delta}_{p_1}^t, \boldsymbol{\delta}_{p_2}^t \in \mathcal{X}^{\text{selected}, t}$
                \STATE $\boldsymbol{\delta}^t_{c_{i}}=\Pi_{[-\epsilon, \epsilon]} [\lambda \boldsymbol{\delta}_{p_1}^t+(1-\lambda) \boldsymbol{\delta}_{p_2}^t], \quad \lambda \sim \mathcal{U}(0,1)$
            \ENDFOR
            \STATE $\mathcal{X}^{\text{crossover}, t} \gets \{\boldsymbol{x} + \boldsymbol{\delta}_{c_i}^t\}_{i=1}^{\mathcal{N}/3}$
            \STATE Update population: $\mathcal{X}^{t} = \mathcal{X}^{\text{mutated}, t} \cup \mathcal{X}^{\text{crossover}, t} \cup \mathcal{X}^{\text{selected}, t}$
        \ENDFOR

        \STATE \textit{\# Dynamic Loss Weighting}
        \IF{Epoch $\mathbb{T}$ $\leq 2$ }
        \STATE $\alpha = \alpha_{init}, \beta = \beta_{init}$
        \ELSE
        \STATE $w_{\text{acc}}(\mathbb{T}) =\frac{\mathcal{L}_{\text{CE}}^{\mathbb{T}}\left(f_{\boldsymbol{P}}(\boldsymbol{x}), y\right)}{\mathcal{L}_{\text{CE}}^{\mathbb{T}-1}\left(f_{\boldsymbol{P}}(\boldsymbol{x}), y\right)}$
        \STATE $w_{\text{rob}}(\mathbb{T}) =\frac{\mathcal{L}_{\text{KL}}^{\mathbb{T}}\left(f_{\boldsymbol{P}}(\boldsymbol{x}), f_{\boldsymbol{P}}(\boldsymbol{x}+\boldsymbol{\delta}) \right)}{\mathcal{L}_{\text{KL}}^{\mathbb{T}-1}\left(f_{\boldsymbol{P}}(\boldsymbol{x}), f_{\boldsymbol{P}}(\boldsymbol{x}+\boldsymbol{\delta}) \right)}$
        \STATE $\alpha = \alpha_{init} \times 2.0 \times \frac{\exp \left(w_{\text{acc}}(\mathbb{T}) / T\right)}{\exp \left(w_{\text{acc}}(\mathbb{T}) / T\right)+ \exp \left(w_{\text{rob}}(\mathbb{T}) / T\right)} $
        \STATE $\beta = \beta_{init} \times 2.0 \times \frac{\exp \left(w_{\text{rob}}(\mathbb{T}) / T\right)}{\exp \left(w_{\text{acc}}(\mathbb{T}) / T\right)+ \exp \left(w_{\text{rob}}(\mathbb{T}) / T\right)}$

        \ENDIF
        
        \STATE \textit{\# Compute loss and update prompts}
        \STATE $\mathcal{L}=\alpha \mathcal{L}_{\text{CE}}\left(f_{\boldsymbol{P}}(\boldsymbol{x}), y\right) + \beta \mathcal{L}_{\text{KL}}\left(f_{\boldsymbol{P}}(\boldsymbol{x}), f_{\boldsymbol{P}}(\boldsymbol{x}+\boldsymbol{\delta}) \right)$
        \STATE Update prompts: $\boldsymbol{P} \gets \boldsymbol{P} - \alpha \times \nabla_{\boldsymbol{P}} \mathcal{L}$
    \ENDFOR
\ENDFOR
\STATE \textbf{Return:} Optimized prompt vectors $\boldsymbol{P}^* = \boldsymbol{P}$
\end{algorithmic}
\end{algorithm*}

\begin{table*}[htbp]
  \centering
    \setlength{\tabcolsep}{28pt}
      \scalebox{0.55}{
    \begin{tabular}{
    c
    l
      S[table-format=2.2]
      S[table-format=2.2]
      S[table-format=2.2]
      S[table-format=2.2]
      S[table-format=2.2]
    }
    \toprule
   \textbf{Dataset} & \textbf{Method} & \multicolumn{1}{c}{\textbf{1-shot}} & \multicolumn{1}{c}{\textbf{2-shot}} & \multicolumn{1}{c}{\textbf{4-shot}} & \multicolumn{1}{c}{\textbf{8-shot}} & \multicolumn{1}{c}{\textbf{16-shot}} \\
    \midrule
    \multirow{5}[2]{*}{\textbf{Average}} & \textbf{C-AVP~\cite{chen2023visual}} & 32.81 & 32.87 & 34.13 & 34.00 & 33.59 \\
          & \textbf{APT~\cite{li2024one}} & 52.02 & 52.85 & 56.42 & 58.68 & 60.73 \\
          & \textbf{AdvPT~\cite{zhang2024adversarial}} & 25.37 & 26.76 & 28.78 & 33.42 & 36.89\\
          & \textbf{AdvMaPLe~\cite{khattak2023maple}} & 28.22 & 34.18 & 44.05 & 54.65 & 64.24 \\
          & \textbf{AdvVLP~\cite{zhou2024fewshot}} & 28.47 & 37.22 & 46.70 & 56.64 & 58.62 \\
          & \textbf{FAP~\cite{zhou2024fewshot}} & 35.42 & 48.17 & 53.38 & 62.17 & 65.32 \\
          & \textbf{ER-APT} & 36.80 & 50.58 & 58.23 & 66.08 & 66.60 \\
    \midrule
    \multirow{5}[2]{*}{\textbf{ImageNet-1K}} & \textbf{C-AVP~\cite{chen2023visual}} & 46.60 & 46.93 & 49.80 & 46.37 & 46.27 \\
          & \textbf{APT~\cite{li2024one}} & 49.30 & 48.83 & 50.90 & 52.03 & 52.63 \\
          & \textbf{AdvPT~\cite{zhang2024adversarial}} & 20.17 & 22.37 & 23.40 & 24.30 & 24.53 \\
          & \textbf{AdvMaPLe~\cite{khattak2023maple}} & 49.27 & 49.97 & 51.27 & 52.13 & 52.93 \\
          & \textbf{AdvVLP~\cite{zhou2024fewshot}} & 49.00 & 50.53 & 51.30 & 52.83 & 53.23 \\
          & \textbf{FAP~\cite{zhou2024fewshot}} & 49.90 & 48.53 & 51.53 & 52.17 & 52.53 \\
          & \textbf{ER-APT} & 48.53 & 49.80 & 52.47 & 53.90 & 54.07\\

    \midrule
    \multirow{5}[2]{*}{\textbf{Caltech101}} & \textbf{C-AVP~\cite{chen2023visual}} & 85.73 & 91.23 & 90.17 & 90.30 & 90.40 \\
          & \textbf{APT~\cite{li2024one}} & 84.77 & 89.70 & 90.77 & 92.37 & 92.93 \\
          & \textbf{AdvPT~\cite{zhang2024adversarial}} & 62.97 & 66.07 & 64.97 & 68.07 & 68.70 \\
          & \textbf{AdvMaPLe~\cite{khattak2023maple}} & 85.53 & 88.00 & 89.53 & 90.63 & 92.17 \\
          & \textbf{AdvVLP~\cite{zhou2024fewshot}} & 85.43 & 87.60 & 89.37 & 90.17 & 92.37 \\
          & \textbf{FAP~\cite{zhou2024fewshot}} & 83.53 & 87.73 & 87.57 & 89.63 & 91.10 \\
          & \textbf{ER-APT} & 84.30 & 87.93 & 89.23 & 90.47 &  90.90 \\
    \midrule
    \multirow{5}[2]{*}{\textbf{DTD}} & \textbf{C-AVP~\cite{chen2023visual}} & 26.97 & 14.27 & 18.77 & 23.63 & 29.20 \\
          & \textbf{APT~\cite{li2024one}} & 41.67 & 45.57 & 51.33 & 54.43 & 54.50 \\
           & \textbf{AdvPT~\cite{zhang2024adversarial}} & 16.73 & 24.57 & 31.70 & 37.47 & 43.77 \\
          & \textbf{AdvMaPLe~\cite{khattak2023maple}} & 13.63 & 16.53 & 6.43 & 33.20 & 57.93 \\
          & \textbf{AdvVLP~\cite{zhou2024fewshot}} & 15.97 & 18.33 & 22.97 & 51.83 & 57.53 \\
          & \textbf{FAP~\cite{zhou2024fewshot}} & 18.40 & 18.40 & 31.27 & 52.13 & 55.17 \\
          & \textbf{ER-APT} & 19.47 & 20.03 & 32.70 & 54.10 & 57.93 \\
    \midrule
    \multirow{5}[2]{*}{\textbf{EuroSAT}} & \textbf{C-AVP~\cite{chen2023visual}} & 9.87 & 9.83 & 10.57 & 9.87 & 18.13 \\
          & \textbf{APT~\cite{li2024one}} & 40.47 & 40.87 & 25.67 & 24.33 & 33.40 \\
          & \textbf{AdvPT~\cite{zhang2024adversarial}} & 23.27 & 15.77 & 13.23 & 36.57 & 53.33 \\
          & \textbf{AdvMaPLe~\cite{khattak2023maple}} & 15.10 & 21.57 & 29.27 & 27.07 & 54.97 \\
          & \textbf{AdvVLP~\cite{zhou2024fewshot}} & 14.37 & 20.37 & 13.20 & 10.87 & 15.50 \\
          & \textbf{FAP~\cite{zhou2024fewshot}} & 31.37 & 43.80 & 64.37 & 76.57 & 81.70 \\
          & \textbf{ER-APT} & 32.53 & 44.57 & 66.03 & 79.20 & 79.13 \\
    \midrule
    \multirow{5}[2]{*}{\textbf{OxfordPets}} & \textbf{C-AVP~\cite{chen2023visual}} & 57.60 & 47.13 & 57.80 & 57.43 & 56.40 \\
          & \textbf{APT~\cite{li2024one}} & 70.23 & 72.87 & 71.83 & 82.87 & 83.70 \\
          & \textbf{AdvPT~\cite{zhang2024adversarial}} & 37.93 & 39.17 & 44.13 & 44.20 & 46.27 \\
          & \textbf{AdvMaPLe~\cite{khattak2023maple}} & 30.67 & 34.03 & 30.70 & 55.60 & 83.27 \\
          & \textbf{AdvVLP~\cite{zhou2024fewshot}} & 29.63 & 31.27 & 67.43 & 80.67 & 82.93 \\
          & \textbf{FAP~\cite{zhou2024fewshot}} & 49.23 & 64.23 & 42.10 & 79.47 & 81.90 \\
          & \textbf{ER-APT} &51.27 & 65.73 & 62.40 & 82.43 & 83.77 \\
    \midrule
    \multirow{5}[2]{*}{\textbf{FGVCAircraft}} & \textbf{C-AVP~\cite{chen2023visual}} & 1.50 & 5.97 & 6.10 & 4.70 & 1.33 \\
          & \textbf{APT~\cite{li2024one}} & 14.77 & 16.37 & 15.70 & 13.60 & 14.77 \\
          & \textbf{AdvPT~\cite{zhang2024adversarial}} & 5.40 & 7.73 & 7.30 & 9.73 & 10.07 \\
          & \textbf{AdvMaPLe~\cite{khattak2023maple}} & 1.37 & 1.80 & 2.50 & 20.37 & 23.63 \\
          & \textbf{AdvVLP~\cite{zhou2024fewshot}} & 1.90 & 6.70 & 14.07 & 14.70 & 23.27 \\
          & \textbf{FAP~\cite{zhou2024fewshot}} & 2.37 & 9.57 & 19.57 & 21.03 & 23.50 \\
          & \textbf{ER-APT} &4.47 & 16.20 & 20.17 & 23.43 & 25.03\\
    \midrule
    \multirow{5}[2]{*}{\textbf{Food101}} & \textbf{C-AVP~\cite{chen2023visual}} & 24.43 & 1.03 & 22.73 & 1.00 & 1.07 \\
          & \textbf{APT~\cite{li2024one}} & 56.57 & 60.17 & 59.80 & 61.57 & 62.50 \\
          & \textbf{AdvPT~\cite{zhang2024adversarial}} & 13.27 & 11.13 & 15.23 & 16.97 & 18.47 \\
          & \textbf{AdvMaPLe~\cite{khattak2023maple}} & 5.27 & 3.10 & 60.00 & 62.70 & 65.13 \\
          & \textbf{AdvVLP~\cite{zhou2024fewshot}} & 1.07 & 1.53 & 41.50 & 61.73 & 43.30 \\
          & \textbf{FAP~\cite{zhou2024fewshot}} & 31.67 & 56.90 & 59.37 & 61.80 & 64.03 \\
           & \textbf{ER-APT} &32.37 & 58.60 & 75.87 & 83.70 & 65.70\\
    \midrule
    \multirow{5}[2]{*}{\textbf{Flowers102}} & \textbf{C-AVP~\cite{chen2023visual}} & 63.10 & 61.47 & 55.97 & 55.50 & 56.17 \\
          & \textbf{APT~\cite{li2024one}} & 61.97 & 67.17 & 82.40 & 84.00 & 86.63 \\
          & \textbf{AdvPT~\cite{zhang2024adversarial}} & 33.97 & 38.47 & 41.97 & 51.13 & 56.03 \\
          & \textbf{AdvMaPLe~\cite{khattak2023maple}} & 1.40 & 46.17 & 52.20 & 83.10 & 87.87 \\
          & \textbf{AdvVLP~\cite{zhou2024fewshot}} & 19.77 & 62.43 & 51.00 & 83.90 & 87.70 \\
          & \textbf{FAP~\cite{zhou2024fewshot}} & 10.40 & 53.10 & 73.13 & 81.53 & 86.27 \\
            & \textbf{ER-APT} &13.38 & 58.60 & 75.83 & 83.77 & 86.83\\
    \midrule
   \multirow{5}[2]{*}{\textbf{StanfordCars}} & \textbf{C-AVP~\cite{chen2023visual}} & 0.57 & 31.20 & 14.00 & 14.40 & 14.83 \\
          & \textbf{APT~\cite{li2024one}} & 40.40 & 15.57 & 43.37 & 49.43 & 51.90 \\
           & \textbf{AdvPT~\cite{zhang2024adversarial}} &  10.80 & 11.17 & 13.47 & 13.27 & 14.87 \\
          & \textbf{AdvMaPLe~\cite{khattak2023maple}} & 25.80 & 39.93 & 44.60 & 50.53 & 56.17 \\
          & \textbf{AdvVLP~\cite{zhou2024fewshot}} & 35.33 & 40.07 & 45.00 & 50.93 & 56.00 \\
          & \textbf{FAP~\cite{zhou2024fewshot}} & 34.70 & 38.60 & 43.20 & 48.47 & 54.23 \\
        & \textbf{ER-APT} &37.83 & 42.47 & 46.87 & 51.77 & 56.07\\
    \midrule
    \multirow{5}[2]{*}{\textbf{SUN397}} & \textbf{C-AVP~\cite{chen2023visual}} & 41.20 & 50.77 & 48.47 & 52.53 & 54.70 \\
          & \textbf{APT~\cite{li2024one}} & 53.53 & 59.20 & 62.37 & 64.30 & 65.67 \\
          & \textbf{AdvPT~\cite{zhang2024adversarial}} & 27.57 & 28.57 & 29.97 & 32.47 & 33.13 \\
          & \textbf{AdvMaPLe~\cite{khattak2023maple}} & 49.70 & 53.73 & 58.23 & 61.50 & 63.57 \\
          & \textbf{AdvVLP~\cite{zhou2024fewshot}} & 48.83 & 53.77 & 57.90 & 61.33 & 63.90 \\
          & \textbf{FAP~\cite{zhou2024fewshot}} & 49.53 & 54.07 & 56.60 & 60.40 & 62.37 \\
           & \textbf{ER-APT} &51.37 & 56.70 & 60.07 & 61.37 & 63.93\\
    \midrule
    \multirow{5}[2]{*}{\textbf{UCF101}} & \textbf{C-AVP~\cite{chen2023visual}} & 3.37 & 1.73 & 1.07 & 18.27 & 0.97 \\
          & \textbf{APT~\cite{li2024one}} & 58.50 & 65.00 & 66.53 & 66.53 & 69.40 \\
          & \textbf{AdvPT~\cite{zhang2024adversarial}} & 27.03 & 29.37 & 31.17 & 33.43 & 36.60 \\
          & \textbf{AdvMaPLe~\cite{khattak2023maple}} & 32.70 & 21.17 & 59.73 & 64.33 & 68.97 \\
          & \textbf{AdvVLP~\cite{zhou2024fewshot}} & 11.83 & 36.83 & 59.97 & 64.07 & 69.10 \\
          & \textbf{FAP~\cite{zhou2024fewshot}} & 28.50 & 54.93 & 58.50 & 60.70 & 65.70 \\
        & \textbf{ER-APT} &29.30 & 55.73 & 58.90 & 62.70 & 68.90 \\
    \bottomrule
    \end{tabular}%
    }
\caption{\textbf{Natural Accuracy (\%) of detailed adversarial few-shot prompt learning results.} We report the mean of the natural accuracy for baselines and our method under different shot number settings across 11 datasets.}
  \label{tab:few_shot_natural_accuracy_appendix}%
\end{table*}%

\begin{table*}[htbp]
  \centering
    \setlength{\tabcolsep}{28pt}
      \scalebox{0.55}{
    \begin{tabular}{
    c
    l
      S[table-format=2.2]
      S[table-format=2.2]
      S[table-format=2.2]
      S[table-format=2.2]
      S[table-format=2.2]
    }
    \toprule
   \textbf{Dataset} & \textbf{Method} & \multicolumn{1}{c}{\textbf{1-shot}} & \multicolumn{1}{c}{\textbf{2-shot}} & \multicolumn{1}{c}{\textbf{4-shot}} & \multicolumn{1}{c}{\textbf{8-shot}} & \multicolumn{1}{c}{\textbf{16-shot}} \\
    \midrule
   \multirow{5}[2]{*}{\textbf{Average}} & \textbf{C-AVP~\cite{chen2023visual}} & 14.04 & 13.20 & 13.08 & 13.77 & 14.28 \\
          & \textbf{APT~\cite{li2024one}} & 3.75 & 4.33 & 4.55 & 5.71 & 6.42 \\
          & \textbf{AdvPT~\cite{zhang2024adversarial}} & 1.05 & 1.32 & 1.59 & 1.86 & 2.04 \\
          & \textbf{AdvMaPLe~\cite{khattak2023maple}} & 8.58 & 12.36 & 18.07 & 25.78 & 32.98 \\
          & \textbf{AdvVLP~\cite{zhou2024fewshot}} & 9.01 & 14.18 & 18.80 & 26.62 & 30.84 \\
          & \textbf{FAP~\cite{zhou2024fewshot}} & 7.88 & 14.05 & 19.59 & 29.51 & 34.61 \\
          & \textbf{ER-APT} & 10.06 & 17.89 & 25.05 & 33.86 & 36.22 \\
    \midrule
    \multirow{5}[2]{*}{\textbf{ImageNet-1K}} & \textbf{C-AVP~\cite{chen2023visual}} & 11.07 & 10.90 & 11.13 & 11.90 & 12.77 \\
          & \textbf{APT~\cite{li2024one}} & 1.30 & 1.03 & 1.40 & 1.80 & 2.07 \\
          & \textbf{AdvPT~\cite{zhang2024adversarial}} & 0.43 & 0.77 & 1.33 & 0.87 & 1.47 \\
          & \textbf{AdvMaPLe~\cite{khattak2023maple}} & 14.60 & 17.13 & 19.00 & 20.60 & 21.90 \\
          & \textbf{AdvVLP~\cite{zhou2024fewshot}} & 15.53 & 17.50 & 19.37 & 20.97 & 22.10 \\
          & \textbf{FAP~\cite{zhou2024fewshot}} & 15.40 & 17.83 & 19.60 & 21.53 & 22.90 \\
          & \textbf{ER-APT} & 17.43 & 20.07 & 21.33 & 22.73 & 23.77 \\
    \midrule
    \multirow{5}[2]{*}{\textbf{Caltech101}} & \textbf{C-AVP~\cite{chen2023visual}} & 50.33 & 55.23 & 52.50 & 50.33 & 52.60 \\
          & \textbf{APT~\cite{li2024one}} & 26.90 & 31.70 & 26.67 & 30.83 & 30.23 \\
          & \textbf{AdvPT~\cite{zhang2024adversarial}} & 7.60 & 8.33 & 7.30 & 10.10 & 9.63 \\
          & \textbf{AdvMaPLe~\cite{khattak2023maple}} & 48.37 & 56.20 & 59.40 & 63.80 & 68.63 \\
          & \textbf{AdvVLP~\cite{zhou2024fewshot}} & 48.47 & 55.33 & 59.07 & 63.13 & 67.97 \\
          & \textbf{FAP~\cite{zhou2024fewshot}} & 41.13 & 53.90 & 57.33 & 62.50 & 67.33 \\
          & \textbf{ER-APT} & 43.37 & 54.93 & 63.47 & 63.93 & 70.53 \\
    \midrule
    \multirow{5}[2]{*}{\textbf{DTD}} & \textbf{C-AVP~\cite{chen2023visual}} & 12.93 & 6.93 & 9.27 & 11.47 & 13.87 \\
          & \textbf{APT~\cite{li2024one}} & 3.83 & 4.27 & 6.33 & 8.70 & 10.47 \\
          & \textbf{AdvPT~\cite{zhang2024adversarial}} & 2.60 & 2.43 & 4.37 & 4.20 & 5.70 \\
          & \textbf{AdvMaPLe~\cite{khattak2023maple}} & 2.93 & 4.20 & 2.40 & 16.97 & 32.17 \\
          & \textbf{AdvVLP~\cite{zhou2024fewshot}} & 4.77 & 7.17 & 10.33 & 25.77 & 32.73 \\
          & \textbf{FAP~\cite{zhou2024fewshot}} & 2.40 & 4.33 & 8.07 & 25.77 & 31.33 \\
          & \textbf{ER-APT} & 3.67 & 5.03 & 10.47 & 26.73 & 32.03 \\
    \midrule
    \multirow{5}[2]{*}{\textbf{EuroSAT}} & \textbf{C-AVP~\cite{chen2023visual}} & 9.80 & 8.67 & 8.50 & 9.77 & 15.83 \\
          & \textbf{APT~\cite{li2024one}} & 0.30 & 0.17 & 0.27 & 0.17 & 0.87 \\
          & \textbf{AdvPT~\cite{zhang2024adversarial}} & 0.00 & 0.37 & 0.00 & 0.00 & 0.17 \\
          & \textbf{AdvMaPLe~\cite{khattak2023maple}} & 0.57 & 5.37 & 16.13 & 21.60 & 32.97 \\
          & \textbf{AdvVLP~\cite{zhou2024fewshot}} & 0.20 & 6.30 & 6.83 & 12.23 & 17.30 \\
          & \textbf{FAP~\cite{zhou2024fewshot}} & 0.00 & 1.00 & 3.60 & 29.30 & 39.73 \\
          & \textbf{ER-APT} & 0.00 & 4.73 & 7.97 & 30.70 & 42.50  \\
    \midrule
    \multirow{5}[2]{*}{\textbf{OxfordPets}} & \textbf{C-AVP~\cite{chen2023visual}} & 22.73 & 15.10 & 16.20 & 17.33 & 16.43 \\
          & \textbf{APT~\cite{li2024one}} & 0.60 & 1.07 & 2.10 & 3.10 & 4.40 \\
          & \textbf{AdvPT~\cite{zhang2024adversarial}} & 0.13 & 1.17 & 1.73 & 0.27 & 0.23 \\
          & \textbf{AdvMaPLe~\cite{khattak2023maple}} & 4.97 & 6.87 & 9.03 & 21.07 & 36.87 \\
          & \textbf{AdvVLP~\cite{zhou2024fewshot}} & 3.83 & 7.07 & 18.47 & 29.63 & 35.57 \\
          & \textbf{FAP~\cite{zhou2024fewshot}} & 3.47 & 12.67 & 9.30 & 34.57 & 41.00 \\
        & \textbf{ER-APT} & 10.47 & 19.50 & 23.43 & 36.93 & 43.50 \\
    \midrule
    \multirow{5}[2]{*}{\textbf{FGVCAircraft}} & \textbf{C-AVP~\cite{chen2023visual}} & 0.77 & 1.60 & 1.27 & 1.20 & 0.63 \\
          & \textbf{APT~\cite{li2024one}} & 0.10 & 0.13 & 0.67 & 1.03 & 1.27 \\
          & \textbf{AdvPT~\cite{zhang2024adversarial}} & 0.00 & 0.00 & 0.17 & 0.00 & 0.43 \\
          & \textbf{AdvMaPLe~\cite{khattak2023maple}} & 0.07 & 0.73 & 1.07 & 5.53 & 7.33 \\
          & \textbf{AdvVLP~\cite{zhou2024fewshot}} & 0.90 & 2.27 & 3.73 & 4.40 & 8.40 \\
          & \textbf{FAP~\cite{zhou2024fewshot}} & 0.07 & 1.10 & 3.93 & 6.07 & 7.97 \\
          & \textbf{ER-APT} & 1.30 & 2.53 & 5.10 & 7.33 & 8.53 \\
    \midrule
    \multirow{5}[2]{*}{\textbf{Food101}} & \textbf{C-AVP~\cite{chen2023visual}} & 5.23 & 0.10 & 4.57 & 0.83 & 0.80 \\
          & \textbf{APT~\cite{li2024one}} & 0.83 & 0.87 & 1.63 & 2.33 & 2.63 \\
          & \textbf{AdvPT~\cite{zhang2024adversarial}} & 0.00 & 0.17 & 0.37 & 0.10 & 0.73 \\
          & \textbf{AdvMaPLe~\cite{khattak2023maple}} & 0.30 & 0.67 & 14.83 & 20.13 & 25.27 \\
          & \textbf{AdvVLP~\cite{zhou2024fewshot}} & 0.77 & 1.10 & 11.20 & 19.33 & 16.50 \\
          & \textbf{FAP~\cite{zhou2024fewshot}} & 1.43 & 10.53 & 18.37 & 23.20 & 26.67 \\
          & \textbf{ER-APT} & 4.37 & 24.43 & 41.40 & 55.63 & 26.87 \\
    \midrule
  \multirow{5}[2]{*}{\textbf{Flowers102}} & \textbf{C-AVP~\cite{chen2023visual}} & 29.70 & 26.93 & 23.73 & 23.57 & 22.03 \\
          & \textbf{APT~\cite{li2024one}} & 2.10 & 3.10 & 4.23 & 6.00 & 8.97 \\
           & \textbf{AdvPT~\cite{zhang2024adversarial}} & 0.43 & 0.23 & 0.63 & 0.87 & 0.80  \\
          & \textbf{AdvMaPLe~\cite{khattak2023maple}} & 0.10 & 17.00 & 25.37 & 48.80 & 58.70 \\
          & \textbf{AdvVLP~\cite{zhou2024fewshot}} & 6.57 & 25.17 & 25.80 & 50.90 & 58.70 \\
          & \textbf{FAP~\cite{zhou2024fewshot}} & 0.53 & 19.57 & 38.77 & 52.63 & 61.47 \\
          & \textbf{ER-APT} & 3.47 & 24.47 & 40.43 & 54.60 & 64.43 \\
    \midrule
    \multirow{5}[2]{*}{\textbf{StanfordCars}} & \textbf{C-AVP~\cite{chen2023visual}} & 0.33 & 5.07 & 2.93 & 2.80 & 3.57 \\
          & \textbf{APT~\cite{li2024one}} & 0.23 & 0.13 & 0.83 & 1.17 & 1.60 \\
          & \textbf{AdvPT~\cite{zhang2024adversarial}} & 0.00 & 0.00 & 0.73 & 1.27 & 0.33  \\
          & \textbf{AdvMaPLe~\cite{khattak2023maple}} & 2.77 & 5.20 & 8.70 & 12.80 & 17.57 \\
          & \textbf{AdvVLP~\cite{zhou2024fewshot}} & 3.80 & 5.33 & 9.07 & 13.27 & 17.47 \\
          & \textbf{FAP~\cite{zhou2024fewshot}} & 4.83 & 7.27 & 11.17 & 15.10 & 19.23 \\
        & \textbf{ER-APT} & 6.77 & 10.53 & 13.83 & 17.30 & 20.93 \\
    \midrule
    \multirow{5}[2]{*}{\textbf{SUN397}} & \textbf{C-AVP~\cite{chen2023visual}} & 11.10 & 13.57 & 13.03 & 17.30 & 17.63 \\
          & \textbf{APT~\cite{li2024one}} & 1.23 & 2.03 & 2.90 & 3.40 & 3.67 \\
          & \textbf{AdvPT~\cite{zhang2024adversarial}} & 0.37 & 0.77 & 0.40 & 1.33 & 2.37  \\
          & \textbf{AdvMaPLe~\cite{khattak2023maple}} & 12.67 & 16.33 & 21.53 & 26.30 & 29.70 \\
          & \textbf{AdvVLP~\cite{zhou2024fewshot}} & 12.60 & 17.33 & 21.17 & 26.23 & 29.70 \\
          & \textbf{FAP~\cite{zhou2024fewshot}} & 14.93 & 19.30 & 23.20 & 27.23 & 30.27 \\
        & \textbf{ER-APT} & 15.73 & 21.30 & 24.33 & 28.70 & 30.70 \\
    \midrule
    \multirow{5}[2]{*}{\textbf{UCF101}} & \textbf{C-AVP~\cite{chen2023visual}} & 0.40 & 1.07 & 0.80 & 4.93 & 0.93 \\
          & \textbf{APT~\cite{li2024one}} & 3.87 & 3.10 & 3.03 & 4.30 & 4.40 \\
          & \textbf{AdvPT~\cite{zhang2024adversarial}} & 0.00 & 0.23 & 0.47 & 1.40 & 0.53  \\
          & \textbf{AdvMaPLe~\cite{khattak2023maple}} & 7.07 & 6.20 & 21.30 & 25.93 & 31.67 \\
          & \textbf{AdvVLP~\cite{zhou2024fewshot}} & 1.73 & 11.43 & 21.77 & 26.97 & 32.80 \\
          & \textbf{FAP~\cite{zhou2024fewshot}} & 2.43 & 7.03 & 22.13 & 26.67 & 32.80 \\
          & \textbf{ER-APT} & 4.13 & 9.27 & 23.77 & 27.87 & 34.67 \\
    \bottomrule
    \end{tabular}%
    }
\caption{\textbf{Robust Accuracy (\%) of detailed adversarial few-shot prompt learning results.} We report the mean of the natural accuracy for baselines and our method under different shot number settings across 11 datasets.}
  \label{tab:few_shot_robust_accuracy_appendix}%
\end{table*}%

\begin{table*}[htbp]
  \centering
    \scalebox{0.8}{
      \begin{tabular}{cccccccccc}
    \toprule
    \textbf{Dataset} & \textbf{Class} & \multicolumn{1}{c}{\textbf{Metric}} & \textbf{C-AVP~\cite{chen2023visual}} & \textbf{APT~\cite{li2024one}} &  \textbf{AdvPT~\cite{zhang2024adversarial}} & \textbf{AdvMaPLe~\cite{khattak2023maple}} & \textbf{AdvVLP~\cite{zhou2024fewshot}} & \textbf{FAP~\cite{zhou2024fewshot}} & \textbf{ER-APT} \\
    \midrule
    \multirow{4}[4]{*}{\textbf{Average}} & \multirow{2}[2]{*}{\textbf{Base}} & Natural Acc  & 31.68 & 18.21 & 43.87 & 60.38 & 58.95 & 70.52 & 71.76 \\
          &       & Adv Acc & 14.43 & 3.80 & 3.50 & 30.69 & 32.37 & 38.05 & 41.81 \\
\cmidrule{2-10}          & \multirow{2}[2]{*}{\textbf{New}} & Natural Acc  & 30.39 & 13.99 & 44.94 & 46.18 & 46.92 & 49.58 & 53.88 \\
          &       & Adv Acc & 13.36 & 3.07 &  8.84 & 20.25 & 21.61 & 21.86 & 26.29 \\
    \midrule
    \multirow{4}[4]{*}{\textbf{ImageNet-1K}} & \multirow{2}[2]{*}{\textbf{Base}} & Natural Acc  & 49.87  & 24.73 & 26.53 & 58.40 & 58.47 & 58.10 & 58.40 \\
          &       & Adv Acc & 12.27 & 9.83 & 0.50 & 25.33 & 24.93 & 25.83 & 25.87 \\
\cmidrule{2-10}          & \multirow{2}[2]{*}{\textbf{New}} & Natural Acc  & 44.80 & 25.43 & 69.03 & 48.83 & 48.67 & 47.83 & 48.53\\
          &       & Adv Acc & 12.27 & 5.90 & 14.77 & 21.03 & 20.50 & 21.57 & 21.93\\
    \midrule
    \multirow{4}[4]{*}{\textbf{Caltech101}} & \multirow{2}[2]{*}{\textbf{Base}} & Natural Acc  & 92.83 & 67.63 & 72.27 & 94.40 & 94.87 & 94.07 & 94.73 \\
          &       & Adv Acc & 57.17 & 15.97 & 13.60 & 73.90 & 76.23 & 74.20 & 77.93 \\
\cmidrule{2-10}          & \multirow{2}[2]{*}{\textbf{New}} & Natural Acc  & 88.83 & 43.83 & 62.33 & 83.27 & 84.47 & 76.53 & 78.53 \\
          &       & Adv Acc & 49.13 & 9.97 & 15.17 & 56.70 & 57.67 & 50.00 & 55.03 \\
    \midrule
    \multirow{4}[4]{*}{\textbf{DTD}} & \multirow{2}[2]{*}{\textbf{Base}} & Natural Acc  & 23.27 & 14.17 & 52.70 & 43.40 & 48.63 & 69.17 & 67.40 \\
          &       & Adv Acc & 10.03 & 8.87 & 7.13 & 21.50 & 27.57 & 41.63 & 40.77\\
\cmidrule{2-10}          & \multirow{2}[2]{*}{\textbf{New}} & Natural Acc  & 13.23 & 19.43 & 46.77 & 21.27 & 22.87 & 35.17 & 41.30 \\
          &       & Adv Acc & 7.20 & 3.60 & 6.83 & 9.97 & 12.37 & 19.77 & 21.30 \\
    \midrule
    \multirow{4}[4]{*}{\textbf{EuroSAT}} & \multirow{2}[2]{*}{\textbf{Base}} & Natural Acc  & 18.07 & 31.37 & 68.73 & 54.30 & 49.03 & 87.70 & 90.27 \\
          &       & Adv Acc & 17.77 & 1.47 & 3.10 & 15.90 & 38.03 & 51.80 & 61.73 \\
\cmidrule{2-10}          & \multirow{2}[2]{*}{\textbf{New}} & Natural Acc  & 25.50 & 3.30 & 9.13 & 26.73 & 35.63 & 32.80 & 48.83 \\
          &       & Adv Acc & 19.97 & 0.00 & 0.37 & 6.83 & 19.47 & 13.40 & 23.50 \\
    \midrule
    \multirow{4}[4]{*}{\textbf{OxfordPets}} & \multirow{2}[2]{*}{\textbf{Base}} & Natural Acc  & 32.57 & 9.47 & 53.43 & 38.97 & 60.67 & 87.37 & 89.03 \\
          &       & Adv Acc & 12.27 & 0.33 & 1.27 & 16.80 & 31.80 & 34.13 & 48.80 \\
\cmidrule{2-10}          & \multirow{2}[2]{*}{\textbf{New}} & Natural Acc  & 32.30 & 2.73 & 51.17 &  39.67 & 57.90 & 72.13 & 85.90 \\
          &       & Adv Acc & 13.37 & 0.00 & 8.53 & 17.50 & 28.90 & 26.07 & 45.17 \\
    \midrule
    \multirow{4}[4]{*}{\textbf{FGVCAircraft}} & \multirow{2}[2]{*}{\textbf{Base}} & Natural Acc  & 2.30 & 3.20 & 12.17 & 15.00 & 9.93 & 24.83 & 28.17 \\
          &       & Adv Acc & 0.30 & 0.13 & 0.47 & 6.63 & 4.53 & 8.00 & 10.33\\
\cmidrule{2-10}          & \multirow{2}[2]{*}{\textbf{New}} & Natural Acc  & 2.00 & 7.97 & 17.13 &  9.97 & 6.73 & 15.83 & 17.20\\
          &       & Adv Acc & 2.00 & 2.63 & 4.37 & 3.13 & 2.50 & 4.23 & 4.90\\
    \midrule
    \multirow{4}[4]{*}{\textbf{Food101}} & \multirow{2}[2]{*}{\textbf{Base}} & Natural Acc  & 2.27 & 2.97 & 25.07 & 71.37 & 71.40 & 72.37 & 73.63 \\
          &       & Adv Acc & 1.27 & 0.47 & 1.63 & 27.90 & 28.43 & 27.57 & 27.67 \\
\cmidrule{2-10}          & \multirow{2}[2]{*}{\textbf{New}} & Natural Acc  & 2.20 & 8.10 & 53.70&  68.93 & 69.90 & 68.20 & 70.00\\
          &       & Adv Acc & 1.00 & 1.93 & 10.97 & 24.50 & 24.60 & 24.20 & 27.03\\
    \midrule
    \multirow{4}[4]{*}{\textbf{Flowers102}} & \multirow{2}[2]{*}{\textbf{Base}} & Natural Acc  & 50.43 & 2.07 & 70.23 & 88.90 & 56.53 & 89.30 & 88.50 \\
          &       & Adv Acc & 24.63 & 0.13 & 1.17 & 62.80 & 36.70 & 65.50 & 69.27 \\
\cmidrule{2-10}          & \multirow{2}[2]{*}{\textbf{New}} & Natural Acc  & 45.23 & 3.47 & 46.70 & 49.90 & 30.00 & 45.67 & 48.03\\
          &       & Adv Acc & 15.77 & 0.03 & 9.93 & 21.07 & 11.63 & 18.10 & 23.47 \\
    \midrule
    \multirow{4}[4]{*}{\textbf{StanfordCars}} & \multirow{2}[2]{*}{\textbf{Base}} & Natural Acc  & 14.87 & 16.83 & 16.60 & 56.47 & 55.60 & 53.97 & 57.23 \\
          &       & Adv Acc & 2.77 & 1.67 & 3.57 & 16.57 & 16.97 & 18.60 & 23.20 \\
\cmidrule{2-10}          & \multirow{2}[2]{*}{\textbf{New}} & Natural Acc  & 15.53 & 11.17 & 35.57 & 46.03 & 46.00 & 42.67 & 44.63 \\
          &       & Adv Acc & 3.70 & 2.20 & 6.83 & 12.10 & 12.67 & 14.10 & 16.07\\
    \midrule
    \multirow{4}[4]{*}{\textbf{SUN397}} & \multirow{2}[2]{*}{\textbf{Base}} & Natural Acc  & 60.20 & 13.10 & 41.40 & 70.23 & 70.57 & 68.47 & 69.93 \\
          &       & Adv Acc & 18.50 & 0.67&  3.77 & 33.87 & 34.10 & 34.63 & 35.87 \\
\cmidrule{2-10}          & \multirow{2}[2]{*}{\textbf{New}} & Natural Acc  & 62.20 & 11.17 & 59.17 & 63.57 & 63.27 & 61.47 & 62.53 \\
          &       & Adv Acc & 21.10 & 2.23 & 12.83 & 29.83 & 29.40 & 30.77 & 30.67 \\
    \midrule
    \multirow{4}[4]{*}{\textbf{UCF101}} & \multirow{2}[2]{*}{\textbf{Base}} & Natural Acc  & 1.77 & 14.73 & 43.47 & 72.77 & 72.80 & 70.37 & 72.07 \\
          &       & Adv Acc & 1.73 & 2.03 & 0.63 & 36.37 & 36.77 & 36.63 & 38.50 \\
\cmidrule{2-10}          & \multirow{2}[2]{*}{\textbf{New}} & Natural Acc  & 2.47 & 17.37  & 43.60 & 49.83 & 50.70 & 47.10 & 47.23 \\
          &       & Adv Acc & 1.43 & 5.33  & 6.60 & 20.13 & 18.00 & 18.30 & 20.07 \\
        \bottomrule
    \end{tabular}%
    }
\caption{\textbf{Detailed results for base-to-new generalization on 11 datasets.} We report the Natural and PGD-100 Accuracy (\%) on the base and new classes that adapted with 16-shot adversarial prompt learning.}
  \label{tab:base2new_appendix}%
\end{table*}%

\end{document}